\documentclass[10pt,twocolumn,letterpaper]{article}

\usepackage[pagenumbers]{cvpr}              
\usepackage{makecell}
\usepackage{bbding}
\usepackage{xcolor}
\usepackage{pifont}
\newcommand{\cmark}{\textcolor{green!70!black}{\ding{51}}} %
\newcommand{\xmark}{\textcolor{red!80!black}{\ding{55}}}   %
\newcommand{\omark}{\textcolor{cyan!90!black}{\Large\textbf{$\circ$}}}

\usepackage{amsmath,amsfonts,bm}

\def\eqref#1{equation~\ref{#1}}

\def\1{\bm{1}}

\DeclareMathAlphabet{\mathsfit}{\encodingdefault}{\sfdefault}{m}{sl}
\SetMathAlphabet{\mathsfit}{bold}{\encodingdefault}{\sfdefault}{bx}{n}

\definecolor{cvprblue}{rgb}{0.21,0.49,0.74}
\usepackage[pagebackref,breaklinks,colorlinks,allcolors=cvprblue]{hyperref}

\def\paperID{10380} 
\def\confName{CVPR}
\def\confYear{2026}

\title{POLAR: A Portrait OLAT Dataset and Generative Framework for Illumination-Aware Face Modeling}

\author{Zhuo Chen$^{1,2\dagger}$\quad Chengqun Yang$^{1\dagger}$\quad Zhuo Su$^2$\textsuperscript{*}\quad Zheng Lv$^2$\quad Jingnan Gao$^1$\\ Xiaoyuan Zhang$^2$\quad Xiaokang Yang$^1$\quad Yichao Yan$^1$\textsuperscript{*}
\and
$^1$Shanghai Jiao Tong University \quad
$^2$PICO\\
{\tt\small \{ningci5252, ycq0191, gjn0310, xkyang, yanyichao\}@sjtu.edu.cn},
\\ \tt\small{suzhuo13@gamil.com}, \tt\small{\{lvzheng.101, zhangxiaoyuan.1001\}@bytedance.com}
}

\begin{document}
\thispagestyle{plain}  
\twocolumn[{
\renewcommand\twocolumn[1][]{#1}
    \maketitle
    \vspace{-2.5em}
    \begin{center}
        \includegraphics[width=\linewidth]{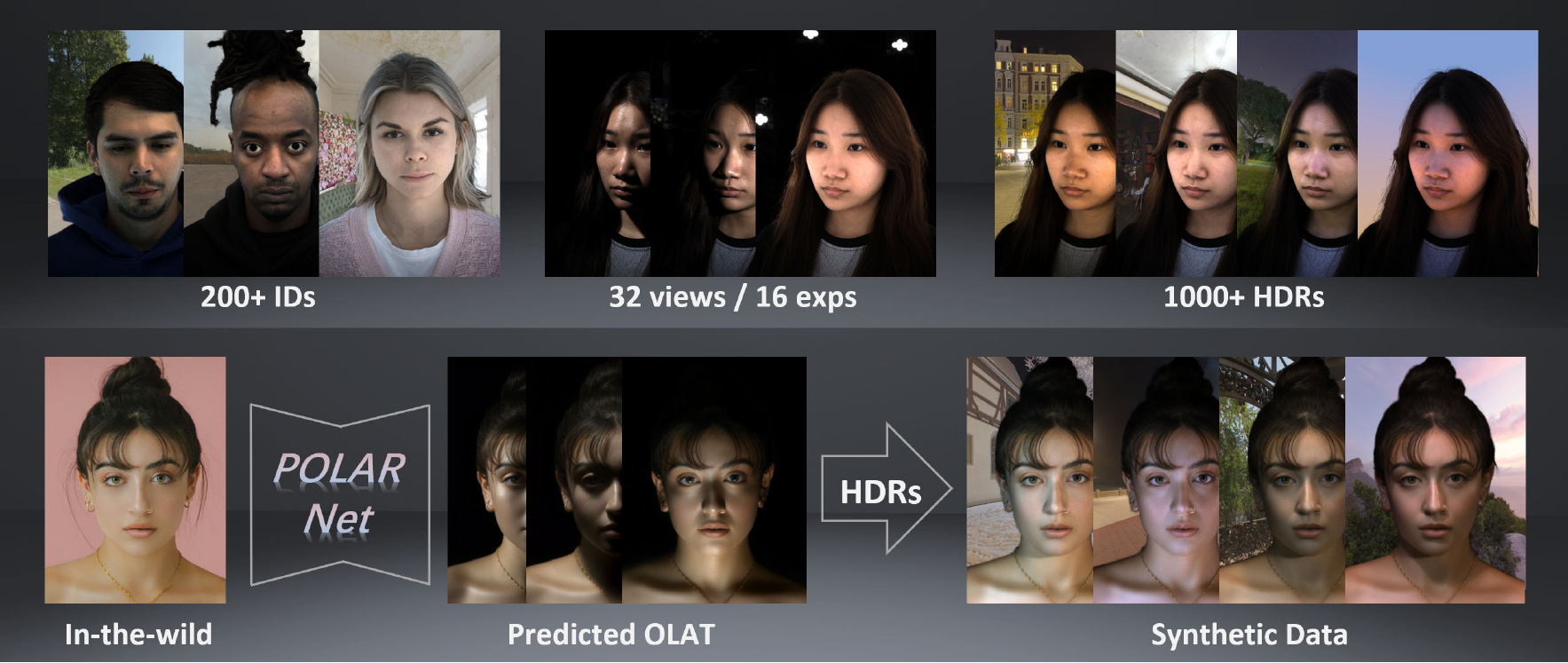}
        \captionof{figure}{
        \textbf{POLAR} captures high-resolution OLAT facial data with diverse subjects and expressions, from which we synthesize large-scale HDR-relit portraits. \textbf{POLARNet} further learns to generate per-light OLAT responses from a single portrait, enabling scalable and physically consistent relighting under arbitrary HDR environments.  Our project page: \href{https://rex0191.github.io/POLAR/}{https://rex0191.github.io/POLAR/}.}
        \label{fig:teaser}
    \end{center}

}]
\pagestyle{plain} 
\footnotetext[1]{{*} Corresponding author, $\dagger$ Equal contribution}
\begin{abstract}

\vspace{-1.0em}
Face relighting aims to synthesize realistic portraits under novel illumination while preserving identity and geometry. 
However, progress remains constrained by the limited availability of large-scale, physically consistent illumination data. 
To address this, we introduce \textbf{POLAR}, a large-scale and physically calibrated One-Light-at-a-Time (OLAT) dataset containing over 200 subjects captured under 156 lighting directions, multiple views, and diverse expressions. 
Building upon POLAR, we develop a flow-based generative model \textbf{POLARNet} that predicts per-light OLAT responses from a single portrait, capturing fine-grained and direction-aware illumination effects while preserving facial identity. 
Unlike diffusion or background-conditioned methods that rely on statistical or contextual cues, our formulation models illumination as a continuous, physically interpretable transformation between lighting states, enabling scalable and controllable relighting. Together, POLAR and POLARNet form a unified illumination learning framework that links real data, generative synthesis, and physically grounded relighting, establishing a self-sustaining “chicken-and-egg’’ cycle for scalable and reproducible portrait illumination.
\end{abstract}

\section{Introduction}
\label{sec:intro}

Face relighting is a long-standing problem in computer vision and computer graphics, with broad applications in photography, cinematography, virtual production, and digital humans. 
Illumination plays a central role in visual perception and storytelling, shaping how light and shadow convey emotion and intent in portrait imagery. 
Although neural rendering~\cite{saito2024rgca} and diffusion-based generation~\cite{liu2025dreamlight, zhang2025scaling, he2024diffrelight} have greatly advanced the realism of portrait relighting, progress is ultimately limited by the scarcity of large-scale, physically consistent illumination data. 
Robust relighting models depend not only on network design but also on the diversity and physical fidelity of the data used for training.

Among various illumination data, \textbf{One-Light-at-a-Time (OLAT)}~\cite{debevec2000acquiring} captures provide the most faithful measurement of facial light transport.
Each capture records the subject’s response to a single directional light, and since light transport is linear with illumination, arbitrary lighting can be reconstructed by linearly combining these single-light bases.
This property makes OLAT a bridge between physics-based rendering and data-driven learning.
However, existing OLAT datasets remain limited in both scale and accessibility.
Most light-stage datasets~\cite{pandey2021total, mei2024holo, mei2025lux, mei2023lightpainter,chaturvedi2025synthlight, he2024diffrelight} used in production are closed-source, while academic releases~\cite{zhang2021neural, saito2024rgca, stratou2011effect} typically cover only a few subjects or expressions and are constrained to low-resolution settings.
As a result, there is still a substantial gap between the growing need for high-fidelity, relightable facial data and the scarcity of publicly available resources.

To fill this gap, we introduce \textbf{POLAR} (Portrait OLAT for Relighting), a large-scale and physically calibrated open dataset for illumination-controlled facial capture.
It contains over 200 subjects recorded under 156 calibrated lighting directions, 32 camera views, and 16 expressions, providing the most diverse portrait OLAT resource available to the open-source community.
In addition to raw captures, POLAR offers relit portraits synthesized under diverse HDR environments, forming a comprehensive platform for data-driven illumination research.
POLAR thus serves as a reliable and physically calibrated resource for developing and validating light transport models.

Although POLAR captures accurately model light transport through physically grounded single-light responses, they are inherently limited to the captured subjects, since each sequence must be recorded under controlled lightstage conditions and cannot generalize to arbitrary individuals.
To overcome this limitation, we develop a flow-based generative model \textbf{POLARNet} that predicts per-light OLAT responses directly from a single uniformly lit portrait. 
The model learns that illumination variations follow consistent physical patterns rather than random image changes, allowing it to capture fine-grained, direction-aware relighting effects. 
Unlike conventional diffusion models~\cite{liu2025dreamlight, jin2024neural_gaffer, zhang2025scaling} that denoise from Gaussian noise, our formulation learns a continuous trajectory that supports \textbf{one-step} inference between the uniform-light images and its corresponding OLAT responses, ensuring that only illumination varies while facial identity remain consistent. 
Conditioned on light direction, the model enables fast and physically controllable synthesis of single-light responses, which can be linearly recombined into full environment relighting results. 
Compared to background-conditioned approaches, our method explicitly models illumination structure and preserves physical interpretability.
By generating per-light responses, the model provides a scalable and low-cost way to synthesize physically consistent illumination data.

\begin{figure}[t]
    \centering
    \includegraphics[width=\linewidth]{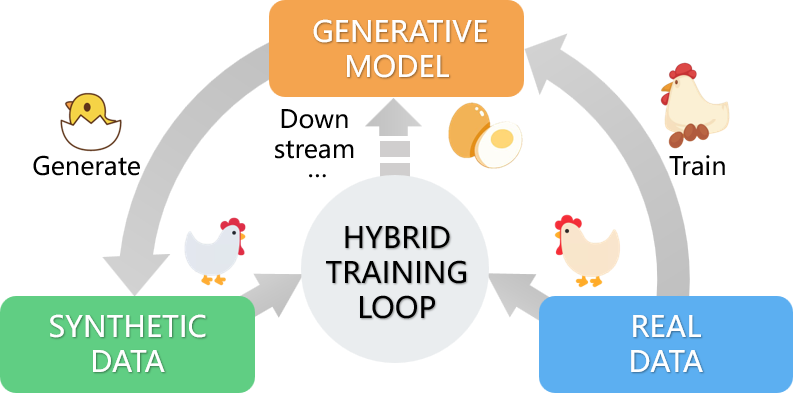}
    \caption{Overview of our “chicken-and-egg” co-evolution loop, where OLAT data guide model learning, and the generative model expands data diversity to benefit downstream tasks.}
    \vspace{-1.5em}
    \label{fig:loop}
\end{figure}
POLAR and its corresponding POLARNet form a unified framework for illumination learning that connects data creation, model training, and physically grounded synthesis. 
As shown in Fig.~\ref{fig:loop}, this interaction forms a self-sustaining ``chicken-and-egg'' cycle, where real data guide the model toward physical accuracy, and the model in turn expands the diversity of illumination data beyond what can be captured in the lab. 
By releasing both the dataset and the model, we aim to promote an open ecosystem for portrait illumination research, supporting not only relighting but also broader downstream tasks such as lighting-aware video generation and digital human rendering.
Our main contributions are summarized as follows:
\begin{itemize}
    \item \textbf{An open, large-scale OLAT dataset.} We present POLAR, a physically calibrated and publicly available OLAT dataset containing 220 subjects, 32 camera views, and 16 expressions, with HDR-relit portraits.
    \item \textbf{A direction-conditioned model for OLAT generation.} We introduce a pair-wise supervised latent bridge formulation that learns physically consistent illumination transitions across light directions, enabling stable and identity-preserving one-step OLAT synthesis.
    \item \textbf{A unified illumination learning framework.} By combining real OLAT captures with generative synthesis, we bridge the gap between costly lightstage acquisition and scalable, physically grounded illumination learning.
\end{itemize}

\section{Related Work}
\label{sec:related}
\subsection{Light Stage}
Light Stage~\cite{debevec2000acquiring} is an active illumination system enabling fine-grained lighting control. This capture system enables collection of detailed multi-view imagery under known illumination, laying the foundation for developing and benchmarking realistic relighting methods~\cite{kim2024switchlight, pandey2021total, saito2024rgca, li2024uravatar, chen2024urhand, guo2019relightables, zhou2023relightable}.
Compared to other setups for capturing real-world multi-illumination data~\cite{carbonera2024relightable, ghosh2011multiview, ma2007rapid}, One-Light-at-a-Time (OLAT) images are facilitated by the Light Stage’s precise, programmable lighting control, enabling acquisition of detailed reflectance data.
Leveraging the linearity of light transport, these OLAT captures can be linearly combined to achieve physically accurate image-based relighting under arbitrary illumination~\cite{peers2007post,debevec2000acquiring, sagar2005reflectance, guo2019relightables}.
A number of Light Stage OLAT datasets for static objects~\cite{toschi2023relight, liu2023openillumination,zhou2025olatverse}, human faces~\cite{stratou2011effect, zhang2021neural, saito2024rgca}, hands~\cite{chen2024urhand}, and full-body humans~\cite{teufel2025humanolat} have been released. However, existing resources are often constrained by limited accessibility or limited diversity, such as small identity counts and lack of expressive variations.
Thus, we collect a fully open, high-quality and diverse OLAT dataset named \textbf{POLAR}, offering a large-scale, high-fidelity foundation for face relighting research. Recently, the concurrent FaceOLAT Dataset~\cite{rao20253dpr} has also been released, highlighting the continued scarcity of open OLAT data and complementing POLAR in scale and accessibility.

\subsection{Portrait Relighting Methods}

Portrait relighting has been extensively studied in both 2D images domain~\cite{zhu2015learning,tsai2017deep,guo2021intrinsic,jiang2021ssh,cong2020dovenet,guo2021image,cong2022high,PIH_wang2023semi,INR,PCT_Guerreiro_2023_CVPR,ren2024relightful, liu2025dreamlight, pandey2021total, kim2024switchlight,sun2019single,wang2020single,kim2022countering,donner2006spectral,zeng2024dilightnet, jin2024neural_gaffer,mei2024holo, mei2023lightpainter, hou2022face, sengupta2021light, yeh2022learning} and 3D avatars~\cite{rao2024lite2relight, sun2021nelf, tan2022volux, bi2021deep, saito2024rgca, zhang2025hravatar, schmidt2025becominglit, sarkar2023litnerf}. 
In the 2D domain, several methods leverage a physics-guided decomposition framework to separate input portraits into intrinsic components and re-render them under HDR environment maps~\cite{pandey2021total, kim2024switchlight, sun2019single, wang2020single}.
While physically interpretable, these approaches often base on simplified BRDF reflectance models, failing to capture complex light transport phenomena such as subsurface scattering~\cite{kim2022countering, donner2006spectral}.
To mitigate reliance on physical priors, recent works adopt diffusion-based models, enabling flexible, photorealistic relighting~\cite{zeng2024dilightnet, jin2024neural_gaffer, ren2024relightful}.
Meanwhile, general relighting frameworks tailored for objects and scenes utilize 3D reconstruction or background-based environmental cues~\cite{xing2025luminet, zhang2025scaling, BiXSMSHHKR2020,Boss2020NeRDNR,Hasselgren2022ShapeLA,jin2023tensoir,kuang2022neroic,munkberg2022extracting,srinivasan2021nerv,yeh2022learning}. However, they remain limited in achieving identity-preserving high-fidelity portrait relighting.
Alternatively, 3D avatar researches focus on constructing relightable human representations that support both environmental lighting and novel viewpoints.
Despite offering strong geometric priors, 3D approaches often rely on parametric templates or specific identities, limiting their generalization and expressiveness~\cite{tewari2021monocular, zhang2020portrait}.
In contrast, \textbf{POLARNet} can generate OLAT data to achieve physically reliable relighting with cross-identity generalization capability.




\begin{figure}[t]
    \centering
    \includegraphics[width=0.95\linewidth]{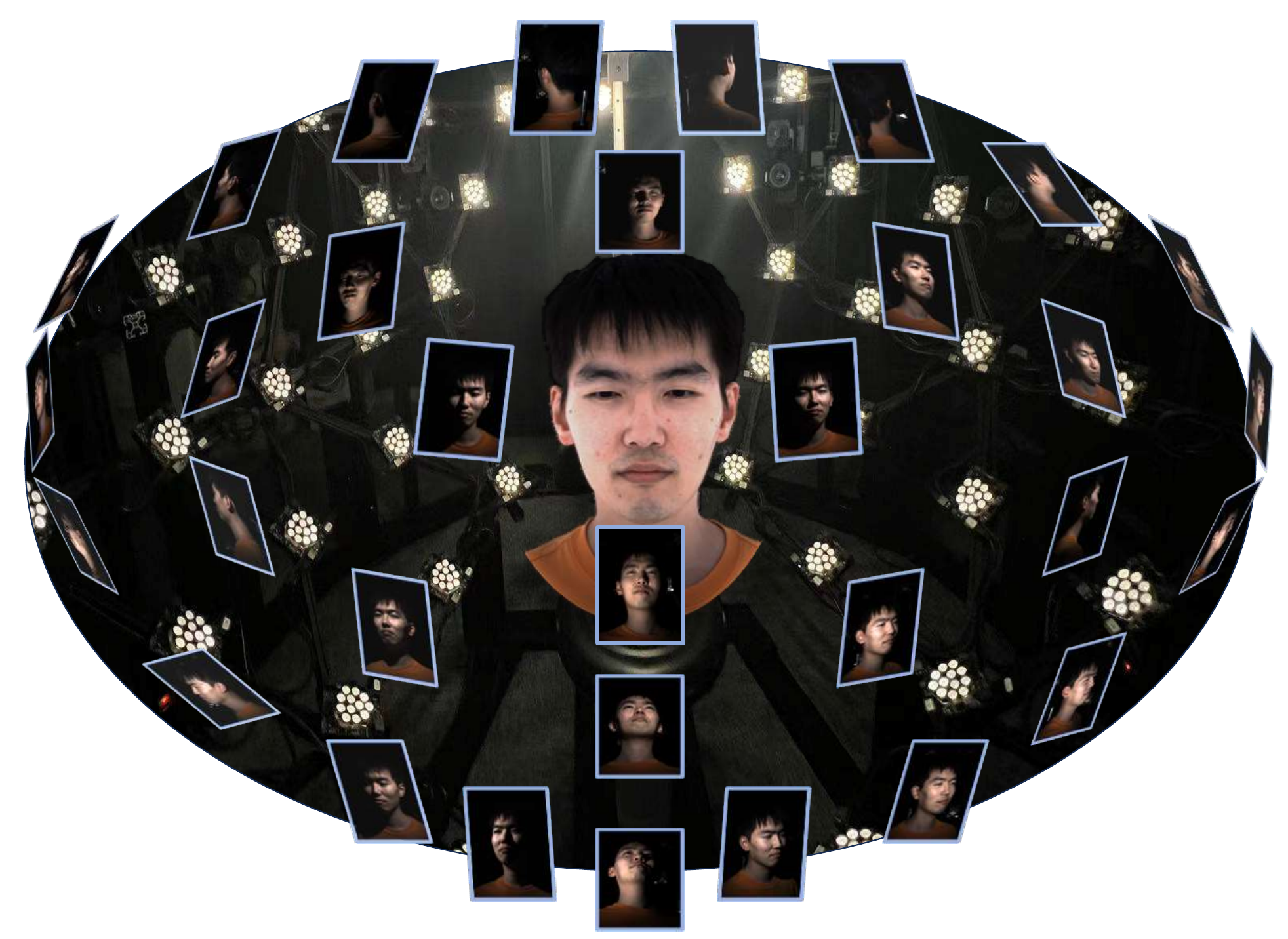}
    \caption{
    Our Light Stage consists of 156 LEDs and 32 synchronized cameras uniformly covering the full sphere. Each light is sequentially activated to produce OLAT captures.}
    \vspace{-0.5em}
    \label{fig:lightsetup}
\end{figure}

\section{The \textit{POLAR} Dataset}
We now introduce \textbf{POLAR}, a large-scale dataset of human faces captured under OLAT illumination. We first describes our acquisition setup (Sec.~\ref{Sec:3_1}) and details the data modalities (Sec.~\ref{Sec:3_2}), and then explains our relighting synthesis pipeline, (Sec.~\ref{Sec:3_3}), and finally compare our POLAR with existing dataset (Sec.~\ref{Sec:3_4}).

\subsection{Acquisition Setup}
\label{Sec:3_1}
We constructed a dedicated Light Stage to acquire high-quality facial OLAT data. 
We employed a multi-view imaging system of \textbf{32} synchronized cameras. 
The Light Stage is equipped with \textbf{156} individually controllable LED light sources, 
distributed in a near-spherical configuration around the subject to cover the full sphere, as shown in Fig.~\ref{fig:lightsetup}. 
Each LED is activated sequentially to record One-Light-at-a-Time (OLAT) captures. 
For each subject, we captured a comprehensive set of \textbf{16 distinct expressions}, 
including neutral, smiling, surprise, and other variations. 
This adds intra-subject diversity and allows analysis of illumination effects under both static and expressive conditions. 
Participants were instructed to avoid glasses, makeup, or reflective accessories to minimize confounding effects. 
In total, our setup captures $\mathbf{157 \times 32 \times 16}$ images per subject (one all-light frame), 
yielding a dense, high-resolution, and multi-view OLAT dataset tailored for face relighting research.
\textbf{Details are included in the supplementary material.}

\begin{figure*}[t]
    \centering
    \includegraphics[width=\linewidth]{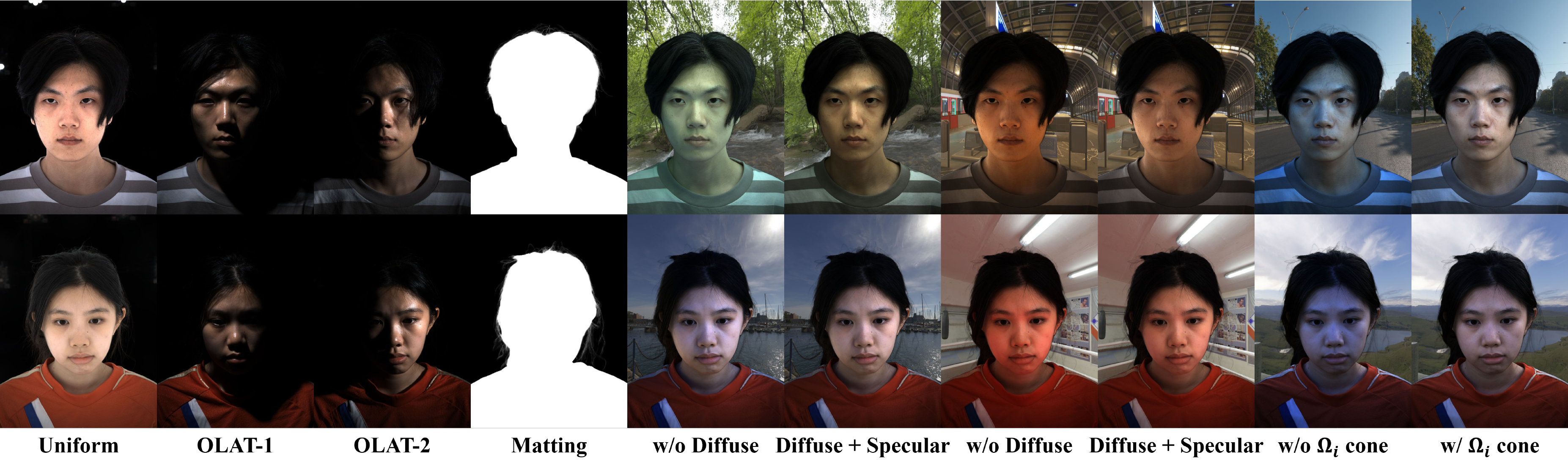}
    \caption{Comparison of OLAT relighting strategies in the POLAR dataset. We evaluate the effect of diffuse–specular separation and light-cone sampling range, showing that both refinements improve shading realism and highlight consistency.}
    \label{fig:relighting}
    \vspace{-1.0em}
\end{figure*}

\subsection{Data Modalities and Annotations}
\label{Sec:3_2}
The POLAR dataset provides multiple modalities supporting both relighting and general face modeling tasks:

\begin{itemize}
    \item \textbf{Raw OLAT captures.} Each subject is recorded under 156 illuminations across 32 views and 16 expressions at 4K resolution.
    \item \textbf{Lighting annotations.} Each OLAT image has calibrated light directions $(\theta,\phi)$ for illumination reconstruction.
    \item \textbf{Alpha mattes.} Per-pixel alpha maps are provided for precise compositing and boundary preservation.
    \item \textbf{Relit portraits.} From captured OLATs, we generate both uniformly lit and HDR-relit portraits. 
    \item \textbf{3D geometry.} Multi-view stereo reconstructions yield per-subject facial meshes aligned with OLAT images.
\end{itemize}

\subsection{Data Processing Pipeline}
\label{Sec:3_3}
\paragraph{Foreground matting.}
High-quality foreground extraction is essential for realistic relighting.
We employ an automatic alpha matting strategy based on the Matte-Anything~\cite{yao2024matte} framework.
To ensure robustness under varying illumination, uniform-light composites are used for matte generation, and mild gamma correction is applied to enhance foreground-background separation. 
\paragraph{{Relighting Synthesis.}}
A central component of POLAR is the generation of relit portraits under arbitrary HDR environment maps, which transforms the raw OLAT captures into a scalable supervision resource. OLAT-based relighting relies on the linear relationship between illumination and appearance. Given single-light captures $\{I_i\}$ and their corresponding light directions $\{l_i\}$, 
the appearance under an environment map $E(\mathbf{l})$ can be approximated by
\begin{equation}
I_{E} \approx \sum_i w_i \, I_i,  
w_i = \int_{\Omega_i} E(\mathbf{l}) \, d\mathbf{l},
\end{equation}
where $E(\mathbf{l})$ denotes the HDR radiance along light direction $\mathbf{l}$ on the unit sphere, 
and $\Omega_i$ denotes the solid-angle region centered at direction $l_i$, corresponding to a 30° illumination cone for each light. 
This linear compositing is widely used and physically sound for diffuse transport,
but specular reflections can introduce exposure imbalance and mild color shifts when simulating colored illuminations.

To enhance the realism of relit results,
we first separate diffuse and specular contributions to prevent global color bias~\cite{kim2024switchlight} (Fig.~\ref{fig:relighting}). Diffuse weights are computed from gray-scale intensity, ensuring illumination color does not overly tint skin tones, while specular weights preserve RGB information to reproduce colored highlights. The final relit image is a balanced combination of the two components:
\begin{equation}
I_E \approx \alpha \sum_i w_i^{\text{diff}} I_i
      + (1-\alpha)\sum_i I_i \odot w_i^{\text{spec}},    
\end{equation}
where $\odot$ denotes element-wise multiplication between RGB specular weights and OLAT images, 
and $\alpha$ is a scene-dependent blending factor. We normalize the accumulated weights to match the total illumination energy of the HDR environment, maintaining consistent global exposure across relit results.
A perceptual tone-mapping is applied to avoid over-saturation under high-intensity illumination, and HDR intensities are calibrated against measured light energy to ensure consistent exposure across relit results. 
We further compare light-cone sampling and point-light approximation in Fig.~\ref{fig:relighting}, where cone-based integration better captures high-intensity local lights and covers full HDR map.

In practice, we compute lighting weights by projecting each HDR environment map onto the 156 calibrated OLAT directions. For each subject, we produce relit portraits under a diverse set of HDR environments, sampled from both procedurally designed “studio-style” setups and publicly available sources such as Poly Haven. 
This synthesis expands each subject’s illumination coverage, producing a diverse set of physically consistent relit portraits that serve as scalable supervision for downstream tasks.

\begin{table*}[t]
\centering
\caption{Comparison of POLAR with existing OLAT-style face datasets. For Accessibility, \cmark and \xmark \ indicate whether the dataset is or is not open source, while $\omark$ means partially open source. POLAR uniquely combines large-scale OLAT captures, calibrated HDR-relit portraits, and full open-source availability, offering a more complete resource for illumination learning.
}
\label{tab:dataset_comparison}
\resizebox{\linewidth}{!}{
\begin{tabular}{lccccccc}
\toprule
Dataset & Subjects & Views & Expressions & Frames & Resolution & Light Types & Accessibility \\
\midrule
Total Relighting~\cite{pandey2021total} & 70 & 6 & 9 & 10.6M & 4K & OLAT + HDR-relit & \xmark  \\
Adobe Data~\cite{mei2024holo, mei2023lightpainter,chaturvedi2025synthlight} & 59 & 4 & 5-15 & 1.2M & 1K & OLAT + HDR-relit & \xmark  \\
NetFlix Data~\cite{mei2025lux,he2024diffrelight} & 67 & 36 & - & 12M & 4K & OLAT + HDR-relit & \xmark  \\
ICT-3DRFE~\cite{stratou2011effect} & 23 & 2 & 15 & 14K & 1K & Gradient / Polarized & \cmark  \\
Dynamic OLAT~\cite{zhang2021neural} & $<10$ & 4 & 1 & 603K & 4K & OLAT & \omark  \\
Goliath-4~\cite{saito2024rgca} & 4 (public) & 100+ & - & $>1$M & Mixed & Relightable captures & \omark   \\
FaceOLAT~\cite{rao20253dpr} & 139 & \textbf{40} & 4 & 5.5M & 4K & OLAT & \cmark \\
\textbf{POLAR (ours)} & \textbf{220} & 32 & \textbf{16} & \textbf{28.8M} & 4K & \textbf{OLAT + HDR-relit} & \cmark \\
\bottomrule
\end{tabular}}
\end{table*}

\noindent\textbf{Background Compositing.}
To enhance realism, each relit portrait is composited with a physically consistent background rendered from the corresponding HDR environment map. The background is precomputed using the subject camera parameters (viewpoint and focal length) to ensure geometric alignment. 

\subsection{Dataset Comparison}
\label{Sec:3_4}
To highlight the contribution of POLAR, we compare it with existing datasets that provide illumination-controlled captures of human subjects in Table~\ref{tab:dataset_comparison}.
Most large-scale illumination datasets developed within industry (e.g., by Adobe~\cite{mei2024holo, mei2023lightpainter,chaturvedi2025synthlight}, NetFlix~\cite{mei2025lux,he2024diffrelight}, or Google~\cite{pandey2021total}) remain proprietary and unavailable to the public due to privacy and licensing constraints. 
Academic efforts have begun to release open OLAT datasets, but most remain limited in scale, diversity, or accessibility.
ICT-3DRFE~\cite{stratou2011effect} contains 23 identities under gradient and polarized lighting, while the Dynamic OLAT dataset~\cite{zhang2021neural} records fewer than ten subjects, limiting its coverage for diverse relighting learning.
Goliath-4~\cite{saito2024rgca} focuses on multi-human or multi-modal assets, yet its full dataset remains under restricted access.
The concurrent FaceOLAT~\cite{rao20253dpr} provides a multi-view face OLAT dataset emphasizing 3D reconstruction and evaluation.
In contrast, our \textbf{POLAR} dataset contains \textbf{220} subjects captured under \textbf{156} calibrated light directions, across \textbf{32} views and \textbf{16} expressions at \textbf{4K} resolution.
Combined with large-scale HDR relighting synthesis, POLAR includes more than \textbf{28.8 million} images in total, offering a substantial increase in both data volume and diversity over existing open OLAT resources.
Along with calibrated meshes, HDR-relit portraits, and a synthesis pipeline, it provides a comprehensive foundation for portrait relighting and illumination-aware modeling.
These features make POLAR one of the first open and large-scale OLAT datasets that bridge the gap between proprietary lightstage data and limited academic releases.

\begin{figure*}[t]
    \centering
    \includegraphics[width=\linewidth]{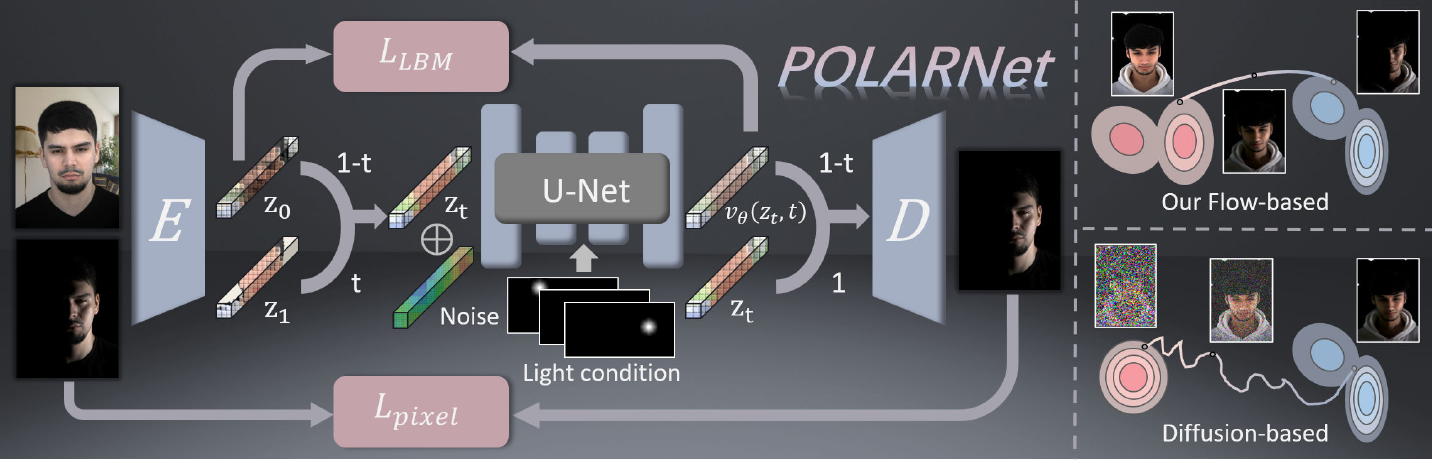}
    \caption{
  \textbf{Overview of our flow-based OLAT data generation framework.} Given a uniformly lit portrait, the encoder–decoder pair $(\mathbf{E},\mathbf{D})$ maps both the input and its target OLAT image into latent space. Latent Bridge Matching learns a continuous, direction-conditioned transport between these endpoints, supervised by the velocity field loss $\mathcal{L}_{\mathrm{LBM}}$. A conditional U-Net predicts the latent drift $v{\theta}(z_t, t, c_{\text{dir}})$ using the encoded light direction as input. During inference, a single forward step transports the latent $z_u$ toward the illumination-specific latent $z_l$, enabling efficient generation of per-light OLAT responses for all calibrated directions. These synthesized OLATs can be linearly composed to render realistic relighting under arbitrary HDR environments.
  }
    \label{fig:pipeline}
\end{figure*}
\section{Flow-based OLAT Generative Model}
Direct lightstage acquisition is costly and inaccessible to most users, and many effective relighting systems remain closed-source. 
Our \textbf{POLAR} dataset bridges this gap by providing a large-scale, publicly available OLAT resource. 
Beyond data release, we further leverage POLAR to train a generative model that learns per-light responses directly from a single uniformly-lit portrait, 
thus enabling low-cost and scalable OLAT-style data creation for arbitrary subjects. The overview is shown in Fig.~\ref{fig:pipeline}.

\subsection{Bridge Matching for Light Transport}
\label{sec:lbm}

\textbf{Motivation.}
Relighting differs from conventional image translation tasks: illumination variations follow physically structured changes rather than random appearance shifts.  
Uniform-light and OLAT images of the same subject share identical appearance, while differing in light direction and intensity. 
However, standard diffusion models learn from noise-to-image mappings and often entangle illumination with texture or identity, leading to inconsistent shading and identity drift.  
To address this, we adopt a \textbf{flow-based latent transport}, which models a continuous trajectory between two semantically aligned endpoints, the uniform light source and the target OLAT image, avoiding random Gaussian initialization.  
This constrains the trajectory to evolve only along the illumination dimension while preserving the intrinsic identity shared by both endpoints.

\paragraph{Pair-wise latent transport.}
Unlike the original Flow Matching~\cite{lipman2022flow} or LBM~\cite{chadebec2025lbm}, which learn a distributional mapping between two unaligned domains,  
we leverage \textit{pair-wise aligned} supervision from OLAT captures, providing deterministic and physically grounded guidance for illumination transport.  
Each training pair $(x_u, x_l^{(\theta,\phi)})$ consists of a uniformly lit portrait and its OLAT portrait illuminated from direction $(\theta,\phi)$.  
We use a lightweight variational autoencoder $(E,D)$, following~\cite{chadebec2025lbm}, to map both images into latent space  
$
z_u = E(x_u)$ and $z_l = E(x_l^{(\theta,\phi)}).
$
The latent interpolation along the bridge is defined as
\begin{equation}
z_t = (1-t)z_u + t z_l + \sigma \sqrt{t(1-t)}\,\epsilon,
\end{equation}
where $\epsilon \!\sim\! \mathcal{N}(0,I)$ introduces limited stochasticity, $\sigma$ controls the noise level and $t\!\in\![0,1]$ parameterizes the illumination path.  
During training, a velocity field $v_\theta(z_t,t,c_{\text{dir}})$ predicts the instantaneous drift that transports $z_t$ toward $z_l$,  
effectively learning a conditional light transport process. 

\paragraph{Direction-conditioned velocity field.}
To achieve controllable illumination, the velocity network $v_\theta(\cdot)$ is conditioned on the light direction  
$c_{\text{dir}} = (\sin\theta, \cos\theta, \sin\phi, \cos\phi)$, represented via sinusoidal positional embeddings.  
The training objective minimizes the discrepancy between the predicted and target drift:
\begin{equation}
\label{eq:lbm}
\mathcal{L}_{\text{LBM}} = 
\mathbb{E}_{t,\epsilon}
\Big[
\|v_\theta(z_t,t,c_{\text{dir}}) - (z_l - z_t)/(1-t)\|_2^2
\Big].
\end{equation}
This formulation encourages the model to capture structured illumination priors and generalize beyond paired data,  
enabling one-step generation of OLAT responses for unseen subjects under arbitrary lighting directions.

\paragraph{Inference.}
Given a uniformly lit portrait $I_{\mathrm{uni}}$ of an arbitrary person, we encode it into the latent space as $z_u = E(I_{\mathrm{uni}})$. 
For a target light direction $\mathbf{L}{=}(\theta,\phi)$, the model directly predicts the transported latent $\hat{z}_l = z_u + (1{-}t)\,v_\theta(z_u, t{=}0, c_{\text{dir}})$ in a \textbf{single forward pass}, 
bypassing the iterative integration typically required in diffusion-based sampling.  
This one-step inference achieves fast and stable generation of OLAT responses while preserving the physical consistency learned during training.  
The decoded output $\hat{I}_{\mathrm{olat}}(\mathbf{L}) = D(\hat{z}_l)$ represents the subject illuminated from direction $\mathbf{L}$, 
and by repeating this process across all calibrated lights, we can synthesize a complete OLAT sequence for the subject.

Once per-light responses are obtained, we can further compose full-environment relighting using the same physically grounded synthesis procedure described in Sec.~\ref{Sec:3_3},  
by linearly integrating the generated OLAT images according to the target HDR environment map.  
The inference and synthesis stages together form a unified, low-cost pipeline for generating realistic portrait illumination under arbitrary lighting conditions.

\begin{figure*}[t]
    \centering
    \includegraphics[width=\linewidth]{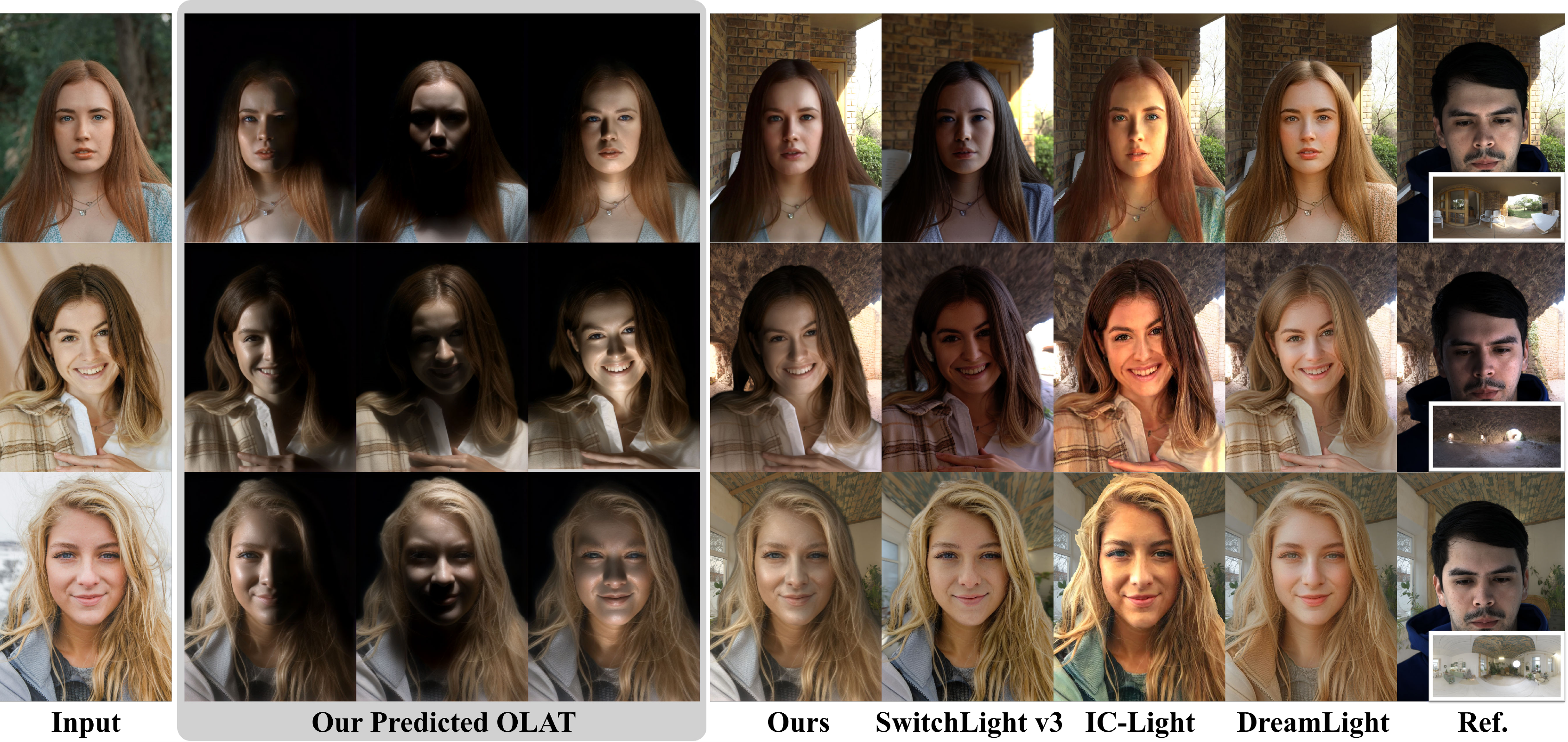}
    \caption{Qualitative comparison of portrait relighting in in-the-wild scenarios. From left to right: input portrait, our generated OLAT responses, relit results from our method, and comparisons with SwitchLight, IC-Light, and DreamLight. The rightmost column shows the reference relit image synthesized from real captured OLAT data.}
    \vspace{-1em}
    \label{fig:Comp_wild}
\end{figure*}

\subsection{Training Details}
\label{training}
Following Sec.~\ref{sec:lbm}, each training pair $(x_0, x_1)$ 
consists of a uniformly lit portrait $I_{\mathrm{uni}}$ and its OLAT counterpart $I_{\mathrm{olat}}(\mathbf{L})$ illuminated from direction $\mathbf{L}{=}(\theta,\phi)$.  
All images are encoded by the pre-trained VAE $(E,D)$ into latent features $(z_0, z_1)$, 
and intermediate samples $z_t$ are constructed via stochastic interpolation.  
The model is optimized using the light-conditioned \textbf{LBM objective} in Eq.~\ref{eq:lbm}.


\noindent\textbf{Identity consistency.}
Since OLAT captures often contain strong shadows, 
we introduce an identity loss computed in image space to preserve facial characteristics:
\begin{equation}
\mathcal{L}_{\text{id}} =
\| f_{\text{id}}(D(z_t)) - f_{\text{id}}(D(z_0)) \|_1,
\end{equation}
where $f_{\text{id}}$ denotes a pre-trained face encoder (ArcFace~\cite{deng2019arcface}) 
and $D(\cdot)$ is the VAE decoder.  

\noindent\textbf{Energy and uncertainty-aware pixel loss.}
To balance photometric contribution across illumination intensities, 
we re-weight the pixel loss using an energy-aware mask, 
where each pixel is scaled by its relative brightness 
$w(x)\!=\!\min(1,\kappa I_{\mathrm{olat}}(x)/\bar{I}_{\mathrm{olat}})$, 
with $\bar{I}_{\mathrm{olat}}$ denoting the mean intensity and $\kappa$ controlling clipping.  
This weighting suppresses dark, low-signal regions and emphasizes well-lit areas, 
resulting in a more perceptually balanced pixel loss:
\begin{equation}
\mathcal{L}_{\mathrm{pix}} =
\|w\odot(\hat{I}_{\mathrm{olat}} - I_{\mathrm{olat}})\|_1.
\end{equation}
To further ensure global exposure consistency and prevent brightness drift between prediction and ground truth, 
we introduce an additional energy regularization term:
\begin{equation}
\mathcal{L}_{\text{energy}} = 
\big\|\tfrac{\|\hat{I}_{\mathrm{olat}}\|_1}{\|I_{\mathrm{olat}}\|_1} - 1\big\|_1.
\end{equation}
Together, these losses encourage physically plausible exposure balance 
and perceptually stable optimization under varying illumination.


\noindent\textbf{Total objective.}
The overall training objective combines the LBM loss with identity, pixel, and energy regularization:
\begin{equation}
\mathcal{L}_{\mathrm{total}} =
\mathcal{L}_{\mathrm{LBM}}
+ \lambda_{\text{id}}\mathcal{L}_{\text{id}}
+ \lambda_{\text{pix}}\mathcal{L}_{\text{pix}}
+ \lambda_{\text{energy}}\mathcal{L}_{\text{energy}}.
\end{equation}


\begin{figure}[t]
    \centering
    \includegraphics[width=\linewidth]{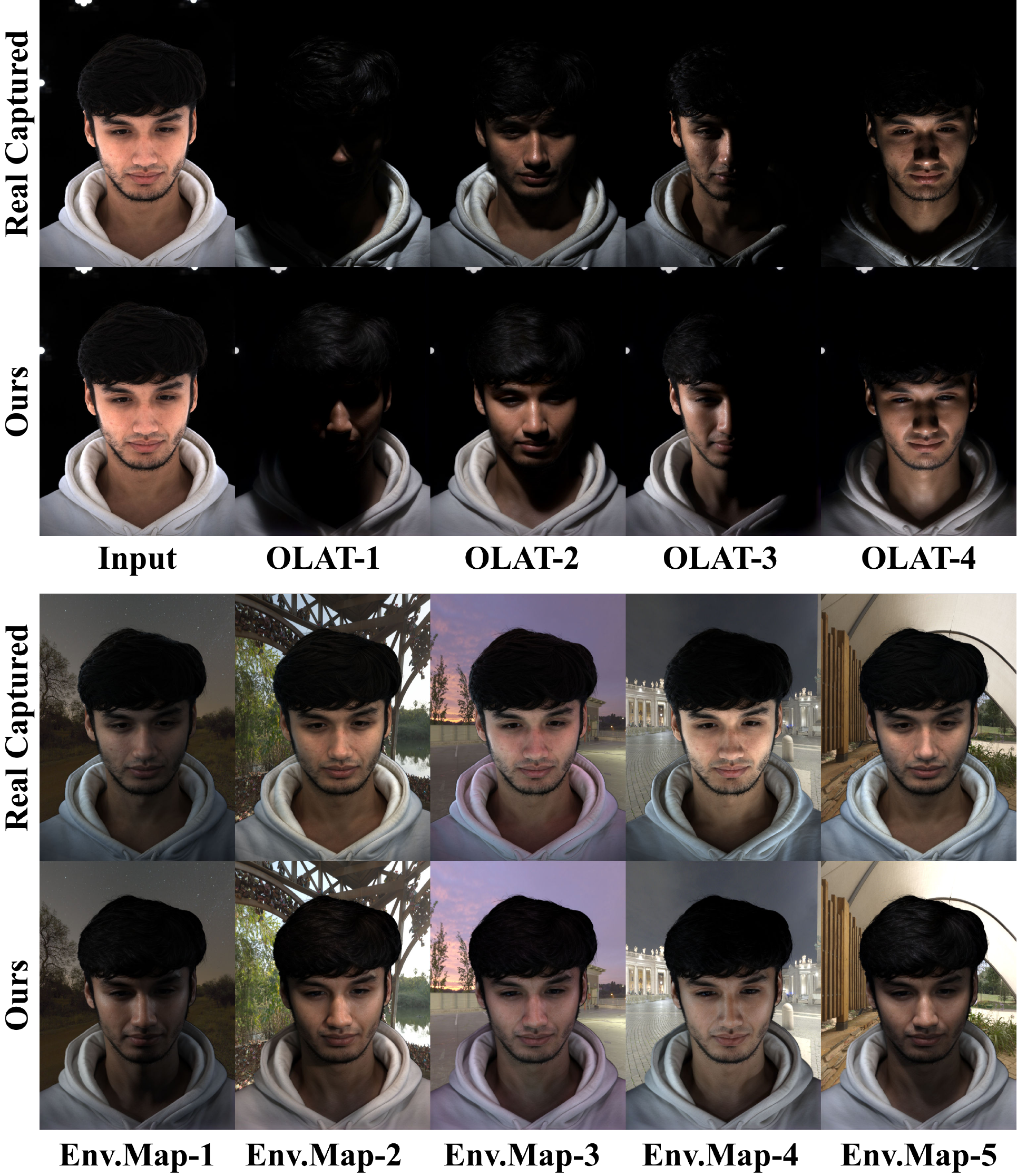}
    \caption{Comparison between generated and captured OLAT data, and their corresponding environment-lit synthesis results. The top part shows OLAT images produced by our model compared with the captured OLAT data. The bottom part presents relit portraits synthesized using the generated OLATs and real OLAT captures under different environment illuminations.}
    \vspace{-1.0em}
    \label{fig:Comp_val}
\end{figure}

\section{Experiments}
\subsection{Comparison of Relit Image Synthesis Pipelines}

To evaluate the fidelity of our OLAT generation model, we compare the synthesized OLAT images with the corresponding captured OLAT data from the POLAR dataset.

\begin{enumerate}
    \item \textbf{Ground-truth OLAT composition.} 
    We directly synthesize relit portraits by linearly combining captured OLAT responses with HDR environment maps. 
    
    \item \textbf{Generated OLAT $\to$ composition.} 
    We first generate a full OLAT sequence for a new subject using our conditional diffusion model, 
    and then compose relit portraits with environment maps. 
\end{enumerate}
As shown in Fig.~\ref{fig:Comp_val}, the generated OLAT results exhibit consistent illumination directionality and realistic shading transitions that closely resemble the captured ground truth.
When these generated OLAT images are further combined to render environment-lit portraits, the resulting relit images align well with those produced from real OLAT captures, demonstrating that our model effectively learns the underlying light transport behavior.

\begin{table}[t]
\caption{Quantitative comparison with state-of-the-art image-based relighting methods.}
\vspace{-0.5em}
\label{tab:quant_comparison}
\centering
\resizebox{0.9\columnwidth}{!}{
\begin{tabular}{lccc}

\hline
Methods & LPIPS$\downarrow$ &  PSNR$\uparrow$ & SSIM$\uparrow$ \\
\hline
SwitchLight~\cite{kim2024switchlight} & 0.168 &  20.69 & \textbf{0.84} \\
IC-Light~\cite{zhang2025scaling} & 0.314 & 18.47 & 0.702 \\
DreamLight~\cite{liu2025dreamlight} & 0.175 & 19.87 & 0.79 \\
Ours & \textbf{0.115} & \textbf{22.12} & 0.82 \\
\hline
\end{tabular}}
\vspace{-1.0em}
\end{table}

\subsection{Baseline Comparisons}
We conduct qualitative comparisons on in-the-wild portraits to assess the visual quality and relighting accuracy between ours and three representative methods.
\textbf{IC-Light}~\cite{zhang2025scaling}, and \textbf{DreamLight}~\cite{liu2025dreamlight} are background-condition relighting methods.
\textbf{SwitchLight}~\cite{kim2024switchlight} adopts a physics-inspired intrinsic decomposition to model relightable portraits.

As illustrated in Fig.~\ref{fig:Comp_wild}, the first three columns show the predicted OLAT images generated by our baseline model from a single input portrait. These results demonstrate that our model can faithfully reconstruct directional lighting responses, including smooth shading and realistic specular highlights.
The following columns present relit results synthesized from arbitrary environment illuminations using the predicted OLAT basis. 
Compared with previous relighting approaches, our method better maintains consistent facial appearance, avoids color drift, and exhibits more natural light transitions.
These qualitative results verify that learning an OLAT decomposition not only enables faithful recovery of light transport but also supports high-quality relighting of in-the-wild portraits.

\subsection{Quantitative Comparisons}

We conduct quantitative evaluations to assess the visual fidelity and perceptual realism of our relighting results. 
Tab.~\ref{tab:quant_comparison} reports the comparison against \textbf{SwitchLight}~\cite{kim2024switchlight}, 
\textbf{IC-Light}~\cite{zhang2025scaling}, and 
\textbf{DreamLight}~\cite{liu2025dreamlight}. 
We use LPIPS$\downarrow$ for perceptual similarity and naturalness, 
and PSNR$\uparrow$ and SSIM$\uparrow$ for pixel-wise reconstruction fidelity. 
All models are evaluated on the test set of relit portraits.
Our method achieves the lowest perceptual error and the highest structural similarity , indicating better preservation of facial details and lighting consistency. 

\begin{figure}[t]
    \centering
    \includegraphics[width=\linewidth]{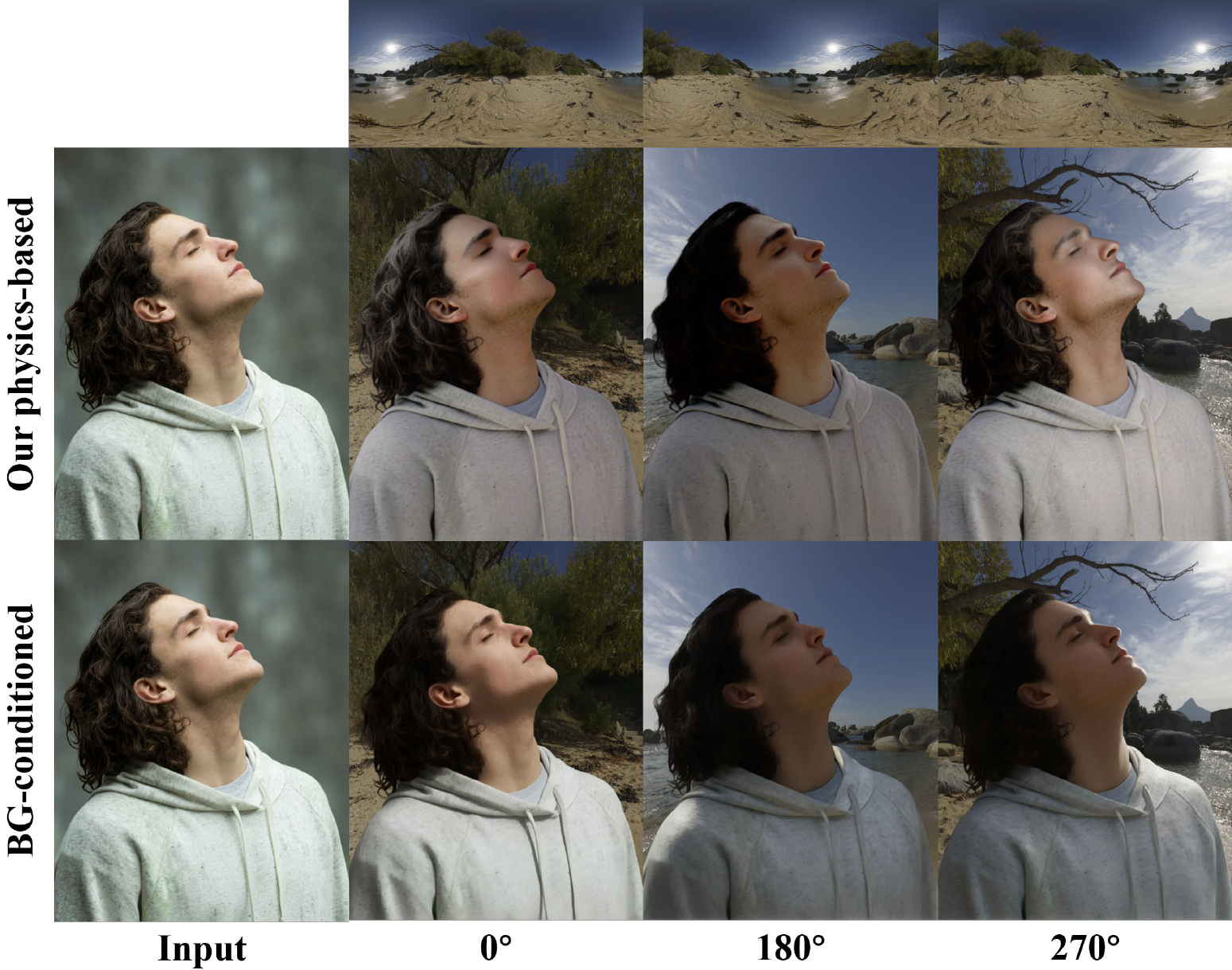}
    \caption{Comparison of relighting consistency under environment rotation. Our Physical OLAT-based relighting preserves consistent shading and highlight movement, while background-conditioned methods show appearance-driven inconsistencies.
    }
    \label{fig:Comp_physics}
    \vspace{-0.5em}
\end{figure}

\begin{figure}[t]
    \centering
    \includegraphics[width=\linewidth]{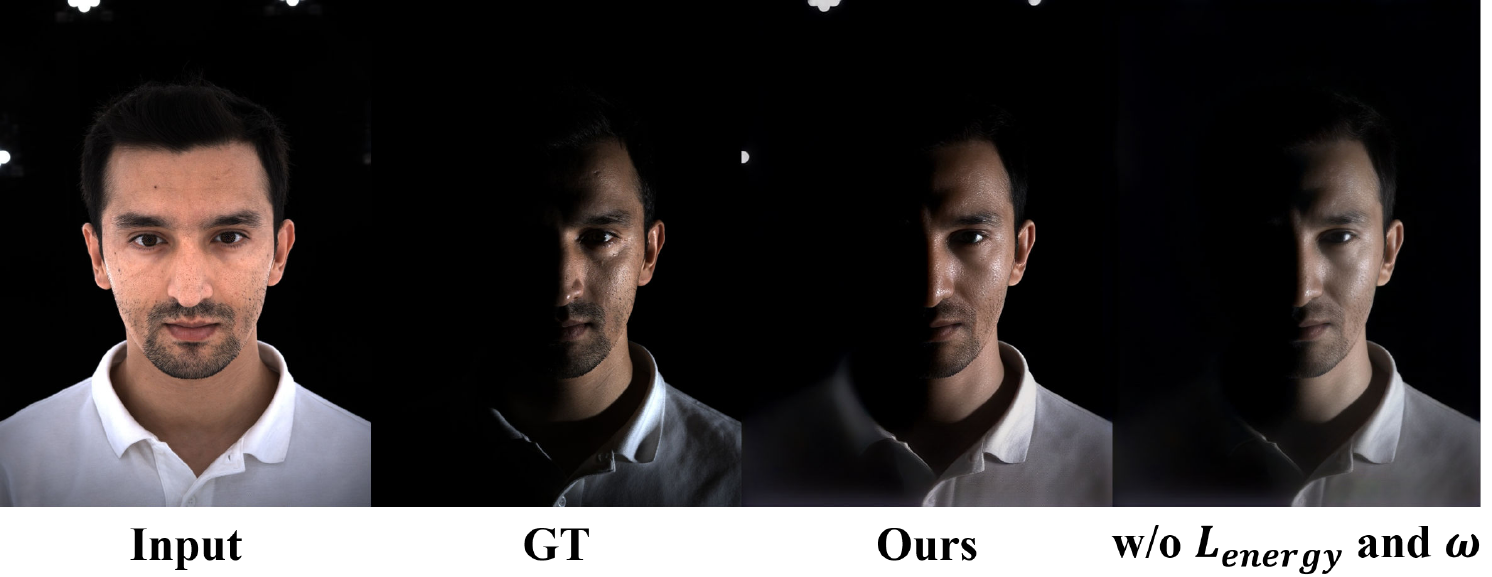}
    \caption{Ablation study on energy and uncertainty-aware loss.
    }
    
    \label{fig:Comp_loss}
    \vspace{-1.0em}
\end{figure}

\subsection{Ablation Study}
\noindent\textbf{Physical consistency under environment rotation.}
We compare our physical OLAT-based relighting with a background-conditioned lighting model~\cite{chadebec2025lbm} under rotating HDR environments (Fig.~\ref{fig:Comp_physics}).
As the HDR map rotates, our method maintains physically consistent shading and highlight movement, consistent with the actual change in lighting direction.
In contrast, the background-conditioned approach relies primarily on contextual cues from the background, leading to inconsistent lighting and shadow discontinuities when the background rotates.

\noindent\textbf{Energy and uncertainty-aware loss.}
We compare models trained with and without the proposed energy-aware terms (Fig.~\ref{fig:Comp_loss}).
Without ${\omega}$ and $\mathcal{L}_{\text{energy}}$, the model tends to overfit low-intensity regions, producing globally darker results and losing contrast in illuminated areas.
By introducing per-pixel weighting $\omega(x)$ that emphasizes bright regions and an energy consistency constraint $\mathcal{L}_{\text{energy}}$, the network learns to preserve overall exposure and recover high-frequency shading details under strong illumination.

\section{Conclusion and Discussion}
We have presented \textbf{POLAR}, a large-scale and physically calibrated portrait dataset offering diverse OLAT captures across identities, expressions, and viewpoints.
Built upon this foundation, our \textbf{POLARNet} learns to generate per-light responses from a single portrait, enabling controllable and physically consistent relighting.
The synthesized OLATs closely approximate real captures, supporting realistic portrait relighting.
This unified dataset–model framework provides a scalable path toward illumination learning and future advances in downstream tasks.

\noindent\textbf{Limitation and future work.}
POLARNet still has several limitations. The generated OLAT data may lose high-frequency details, especially around specular highlights and shadow boundaries, and performance degrades under extreme facial poses or lighting conditions. In future work, we plan to extend the framework to video-based OLAT synthesis for temporally consistent illumination in dynamic faces.

{
    \small
    \bibliographystyle{ieeenat_fullname}
    \bibliography{main}
}
\renewcommand{\thefigure}{\Alph{figure}}
\renewcommand{\thesection}{\Alph{section}}
\setcounter{section}{0}
\setcounter{figure}{0}
\clearpage
\setcounter{page}{1}
\maketitlesupplementary

\begin{figure}[htp]
    \centering
    \includegraphics[width=\linewidth]{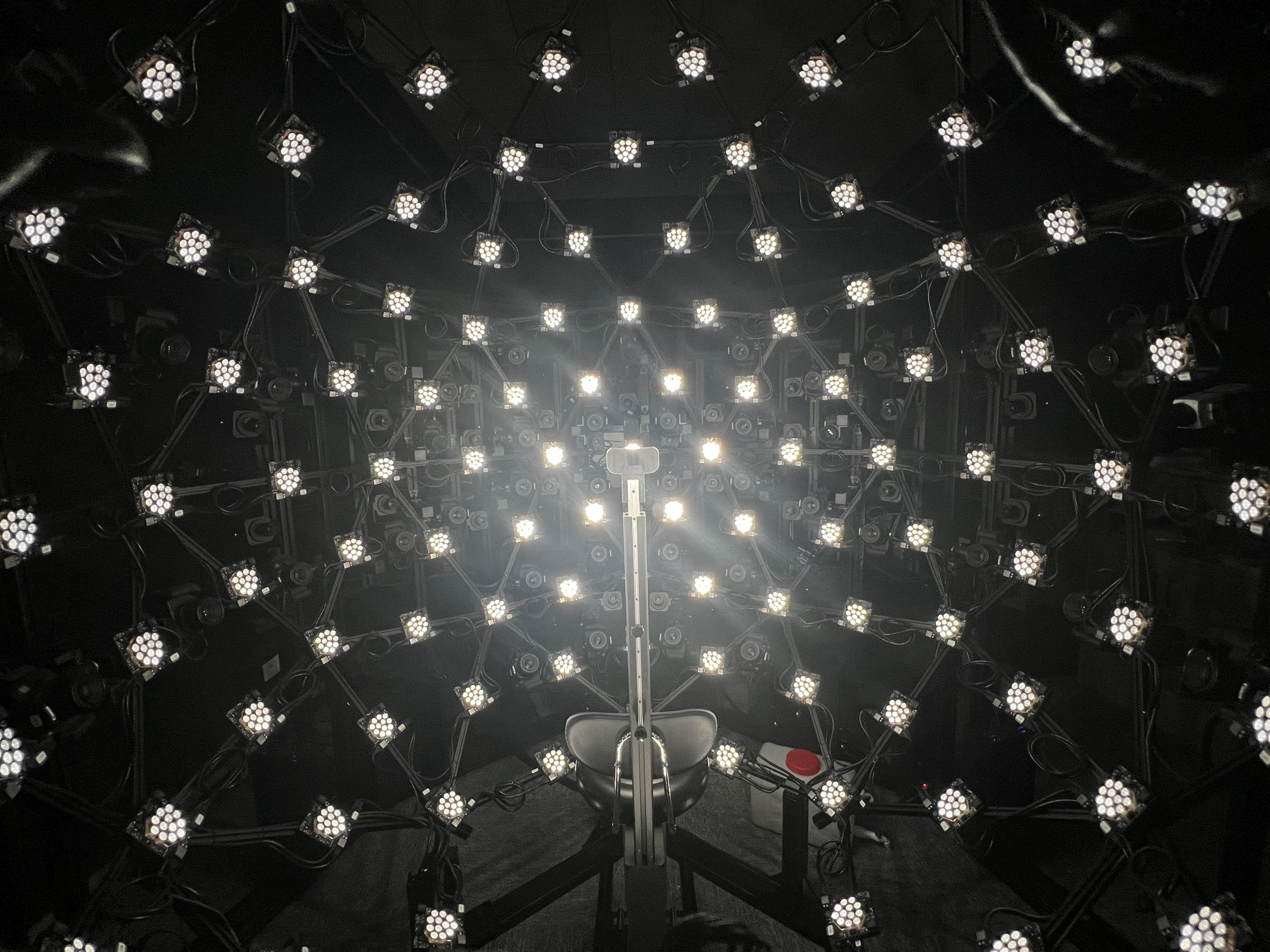}
    \caption{Light stage. \textit{POLAR} is captured using a calibrated light-stage setup that provides controlled illumination and high-fidelity appearance capture.}
    \label{fig:light-stage}
\end{figure}
\section{The \textit{POLAR} Dataset}
\subsection{Acquisition Setup}
We constructed a dedicated Light Stage to acquire high-quality facial OLAT data. 
We employed a multi-view imaging system of \textbf{32} synchronized cameras. 
The Light Stage is equipped with \textbf{156} individually controllable LED light sources, 
distributed in a near-spherical configuration around the subject to cover the full sphere, as shown in Fig.~\ref{fig:light-stage}. 
Each LED is activated sequentially to record One-Light-at-a-Time (OLAT) captures. 
The lights are photometrically calibrated to ensure consistent intensity and color temperature. 
Each unit has a radiation angle of $30^{\circ}$, with illumination designed such that the effective range coincides with the spherical light field radius of the stage. Beyond this range, light intensity drops sharply, ensuring minimal crosstalk between neighboring directions. 
\begin{figure}[t]
    \centering
    \includegraphics[width=\linewidth]{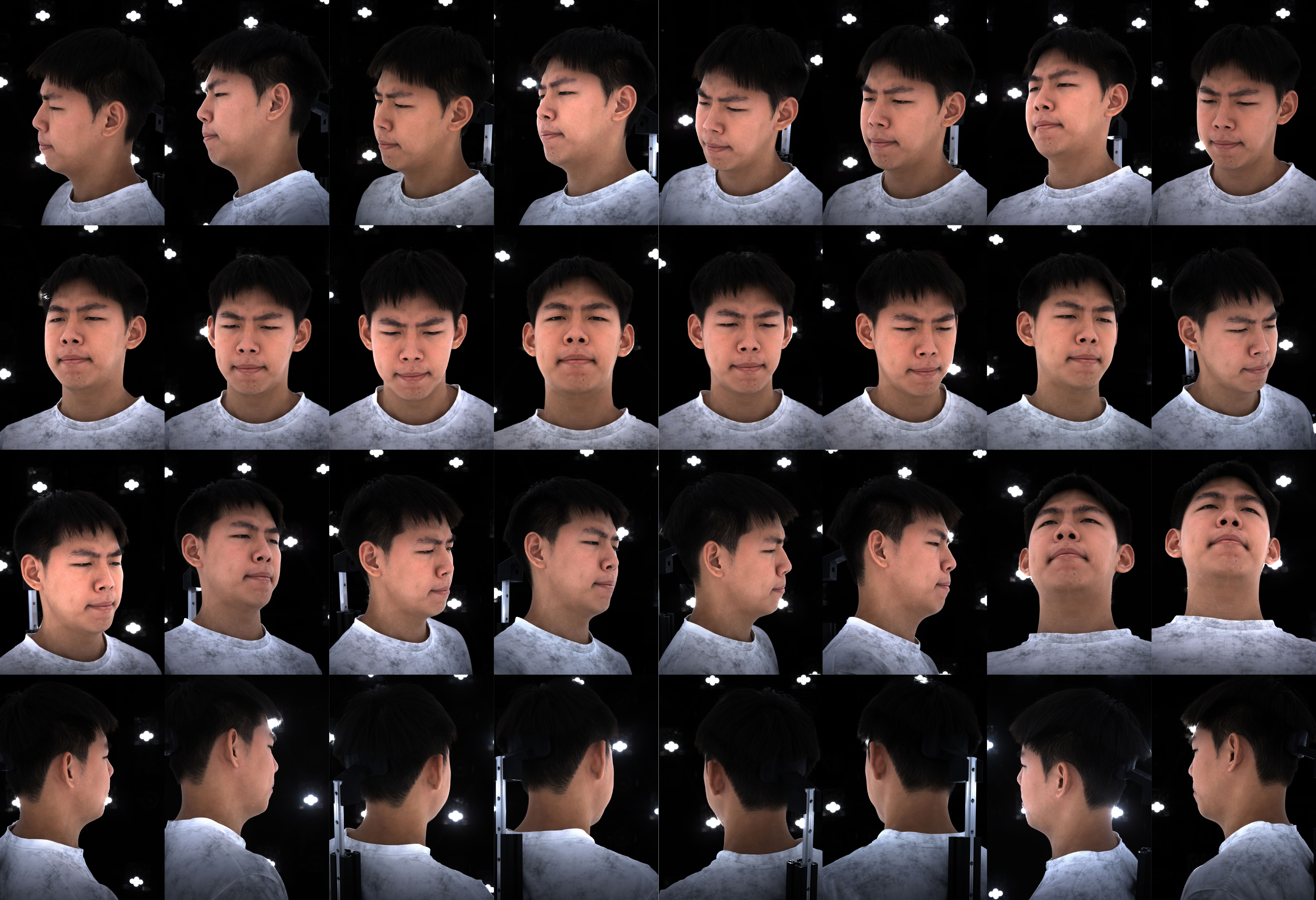}
    \caption{Viewpoints. \textit{POLAR} provides 32 synchronized camera views offering diverse visual perspectives.}
    \label{fig:view}
\end{figure}
\begin{figure}[h]
    \centering
    \includegraphics[width=0.85\linewidth]{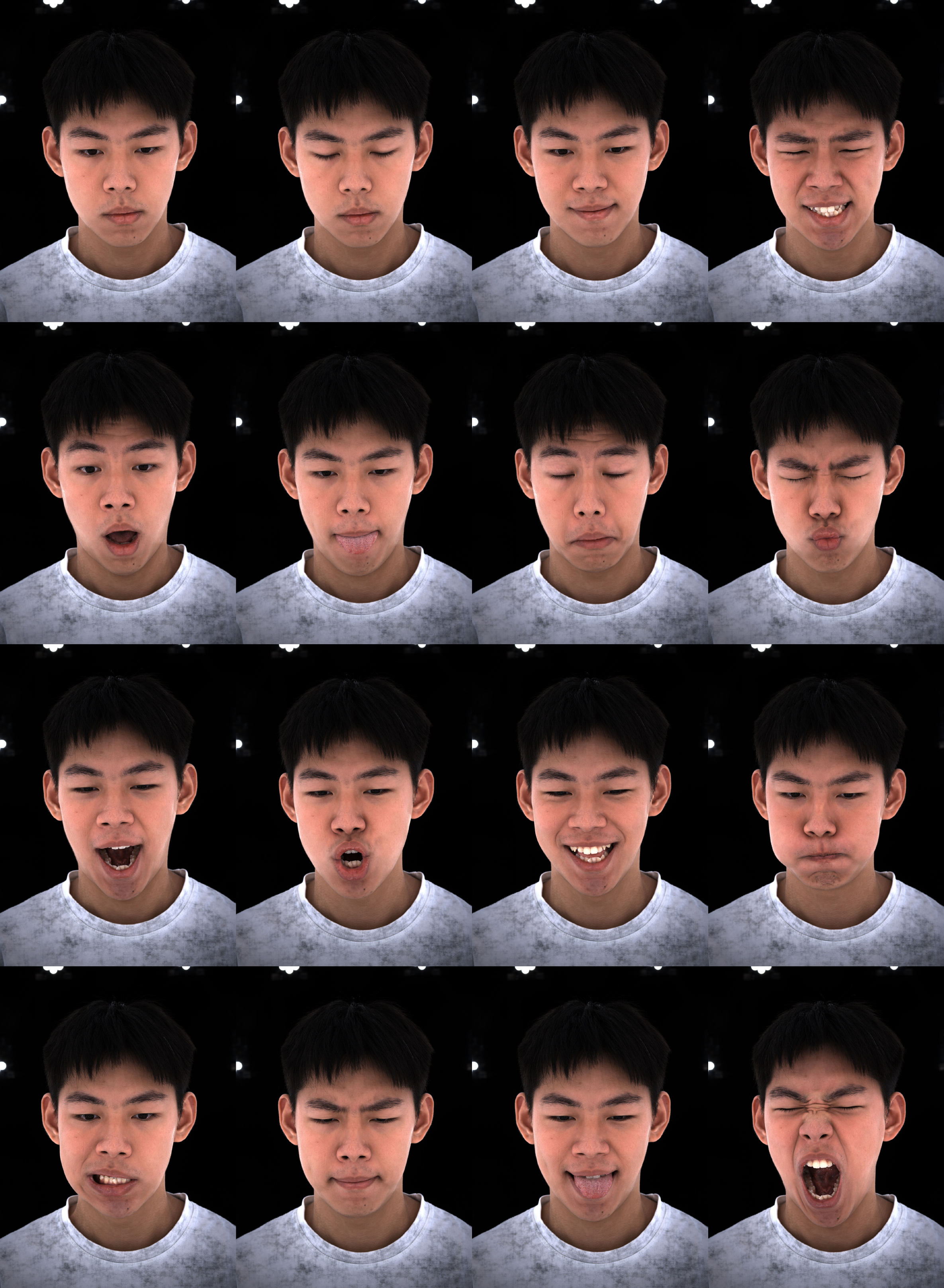}
    \caption{Expressions. \textit{POLAR} includes 16 distinct facial expressions capturing a wide range of appearance variations.}
    \vspace{-1em}
    \label{fig:exp}
\end{figure}
The geometric positions of all light sources are registered in a global spherical coordinate system $(\theta, \phi)$, 
enabling precise annotation of incident illumination. 
The 156 directions provide dense and nearly uniform sampling of the frontal hemisphere (Fig.~\ref{fig:light-stage}). 
Each camera is equipped with a \textbf{35 mm fixed-focal-length lens}, 
\begin{figure*}[t]
    \centering
    \includegraphics[width=\linewidth]{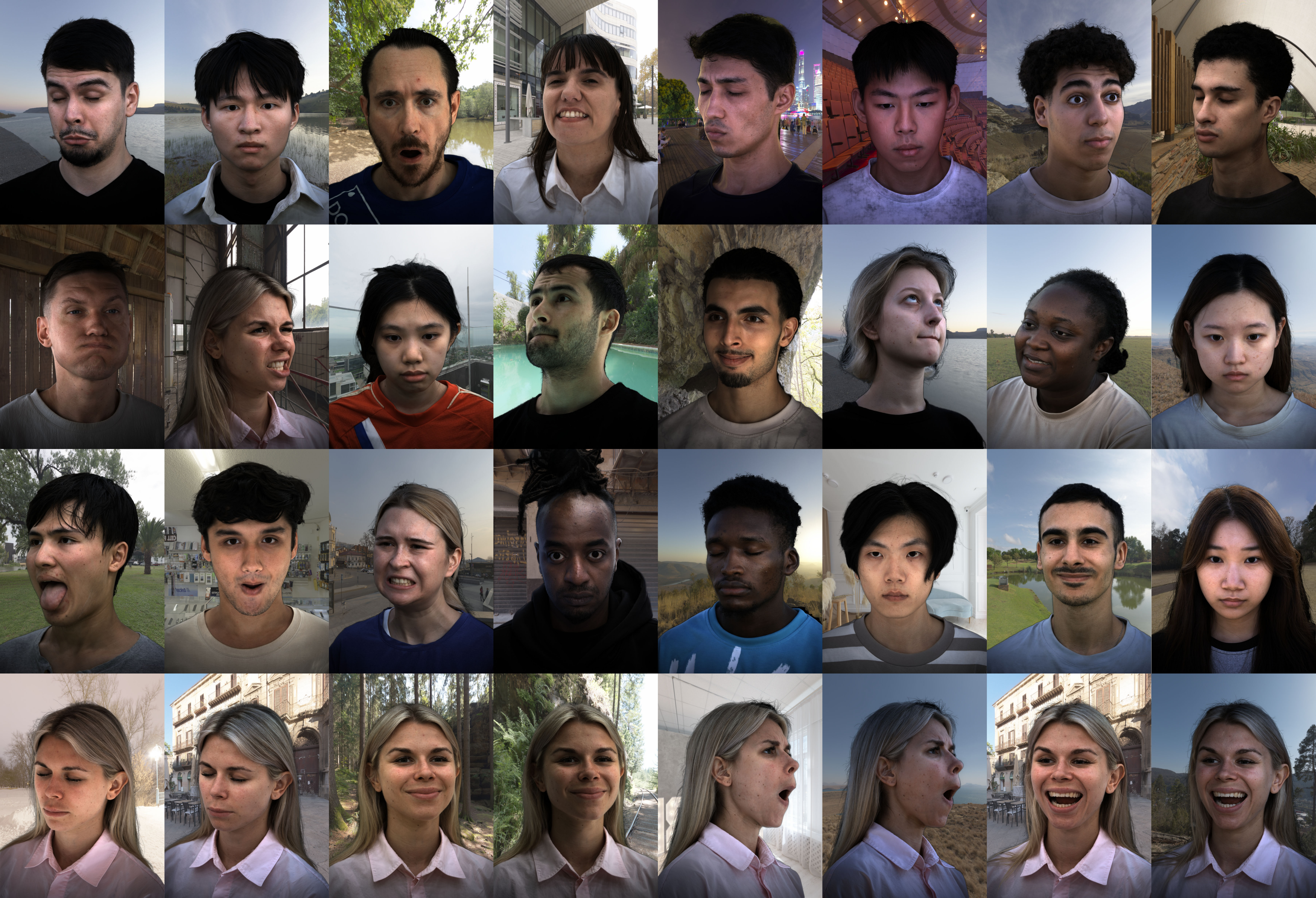}
    \caption{Synthetic relit portraits in our POLAR dataset.}
    \label{fig:olat_hdr}
\end{figure*}
which minimizes geometric distortion while preserving facial proportions. 
We demonstrate portrait images from 32 different viewpoints, as shown in Fig.~\ref{fig:view}.
Images are recorded at \textbf{4K resolution} in linear color space with 16-bit precision, 
retaining both fine-scale details and high dynamic range. 
Multi-view coverage ensures that relighting can be studied not only for frontal images but also across a range of viewpoints, 
supporting applications in 3D reconstruction and view-consistent relighting.

\subsection{Facial Expression}
The real OLAT subset includes 16 controlled facial expressions spanning neutral, mild, moderate, and extreme deformations, as demonstrated in Fig.~\ref{fig:exp}. These expressions cover a wide range of anatomically meaningful configurations, including: neutral and relaxed states; eye-closed and gaze-down variants; lip-compression and cheek-inflation motions; symmetric and asymmetric mouth deformations (closed-mouth smile, open-mouth smile, wide mouth-open, extreme yawning); as well as brow-raising and frowning behaviors. This curated expression set ensures comprehensive coverage of both subtle muscle activations and large-amplitude shape changes.
\begin{figure*}[t]
    \centering
    \includegraphics[width=\linewidth]{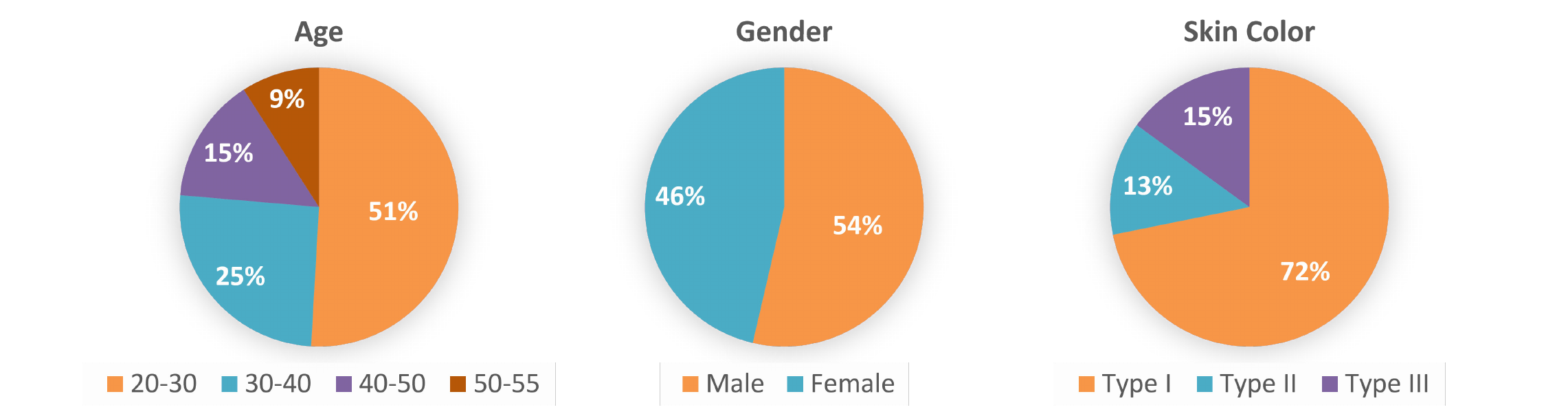}
    \caption{Dataset Summary. We summarize age, gender, skin color in our dataset.}
    \label{fig:data_summary}
\end{figure*}

\subsection{HDR-relit Examples}
To further illustrate the diversity and coverage of our dataset, Fig.~\ref{fig:olat_hdr} presents a representative subset of HDR-relit portraits generated using a wide variety of high-dynamic-range environment maps. The examples span subjects of different skin tones and ethnicities, including Black, White, and Asian individuals, and cover multiple viewing angles as well as a range of facial expressions. Across these variations, the relit results exhibit coherent shading behavior, realistic highlight placement, and consistent shadow geometry under complex outdoor and indoor illumination.
These HDR-relit samples complement the real OLAT captures by exposing each subject to rich, spatially varying light fields that cannot be reproduced with point-light acquisition alone. 
For external HDRs, we provide download scripts and directory listings to ensure reproducibility. 
Through this process, every subject expands from OLAT captures to thousands of HDR-lit portraits under diverse and physically consistent lighting, providing rich supervision for downstream tasks. 

\subsection{Dataset Statistics}
To better characterize the demographic distribution of our OLAT dataset, we provide an analysis across age, gender, and skin‐type attributes, as shown in Fig.~\ref{fig:data_summary}. The dataset primarily consists of young to middle-aged adults, with the majority falling within the 20–30 age range (51\%), followed by 30–40 (25\%), 40–50 (15\%), and 50–55 (9\%). 

Gender distribution is relatively balanced, with 54\% female and 46\% male participants. Such balance helps mitigate gender-related bias and ensures more stable generalization when learning appearance or reflectance priors.

For skin type, the dataset predominantly contains subjects of Type I--III participants. Although the distribution is naturally skewed toward the primary demographic region where data collection was conducted, the inclusion of multiple skin tones improves the dataset’s applicability to cross-ethnicity appearance modeling.

\subsection{Supplementary Data Processing Pipeline}
\subsubsection*{Detailed foreground matting.} 
Accurate foreground extraction is a crucial step in our processing pipeline, as it directly determines the realism of synthesized relit images. In particular, hair strands and semi-transparent boundaries are extremely sensitive to segmentation quality: hard binary masks often produce halo artifacts or clipped silhouettes when composited under new illumination, while high-quality alpha mattes preserve fine details and lead to significantly more natural relighting results. 


We adopt the \textit{Matte-Anything} framework for alpha matting. Since interactive scribbles or clicks are impractical for our large-scale batch processing, we design an automatic initialization strategy: facial keypoints are detected to provide foreground seeds, and a text prompt (\texttt{"person"}) is supplied to guide the model toward the subject region.

As all 156 OLAT images plus the uniform-light portrait are captured with the subject in a fixed pose, we only need to generate a single high-quality matte per subject, which can be reused across lighting conditions. However, directly using the captured uniform-light images is problematic: in some viewpoints, bright light sources appear near the face, leading to segmentation failures. To address this, we instead generate uniform-light images synthetically by averaging OLAT responses, ensuring a consistent background where stage lights do not appear as artifacts.

\begin{figure}[t]
    \centering
    \includegraphics[width=\linewidth]{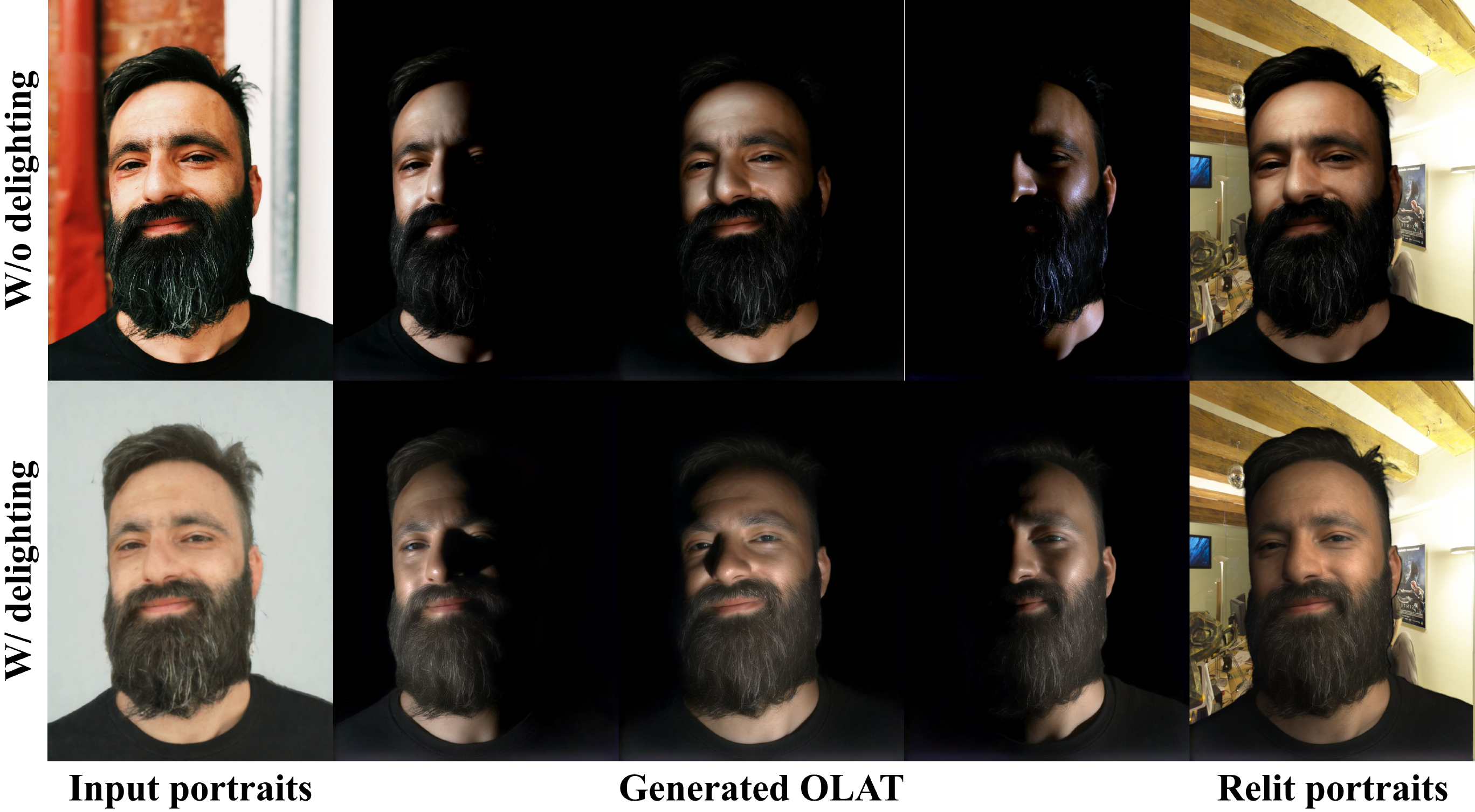}
    \caption{Our failure case and delighting solution. Without delighting, the predicted OLAT images entangle the illumination present in the input, leading to biased shading and inconsistent light responses. After applying the delighting module, the input is normalized to an illumination-neutral appearance, enabling the model to generate more accurate OLAT predictions.}
    \label{fig:failure_case}
\end{figure}

Another challenge arises from the optical setup: because illumination intensity decays sharply beyond the light field radius, the background remains very dark. In this setting, dark regions of hair or clothing often merge with the background, making separation difficult. To mitigate this, we apply strong gamma correction to the uniform-light composites before matting, which enhances contrast and improves foreground-background discrimination in low-light regions.

\begin{figure*}[t]
    \centering
    \includegraphics[width=\linewidth]{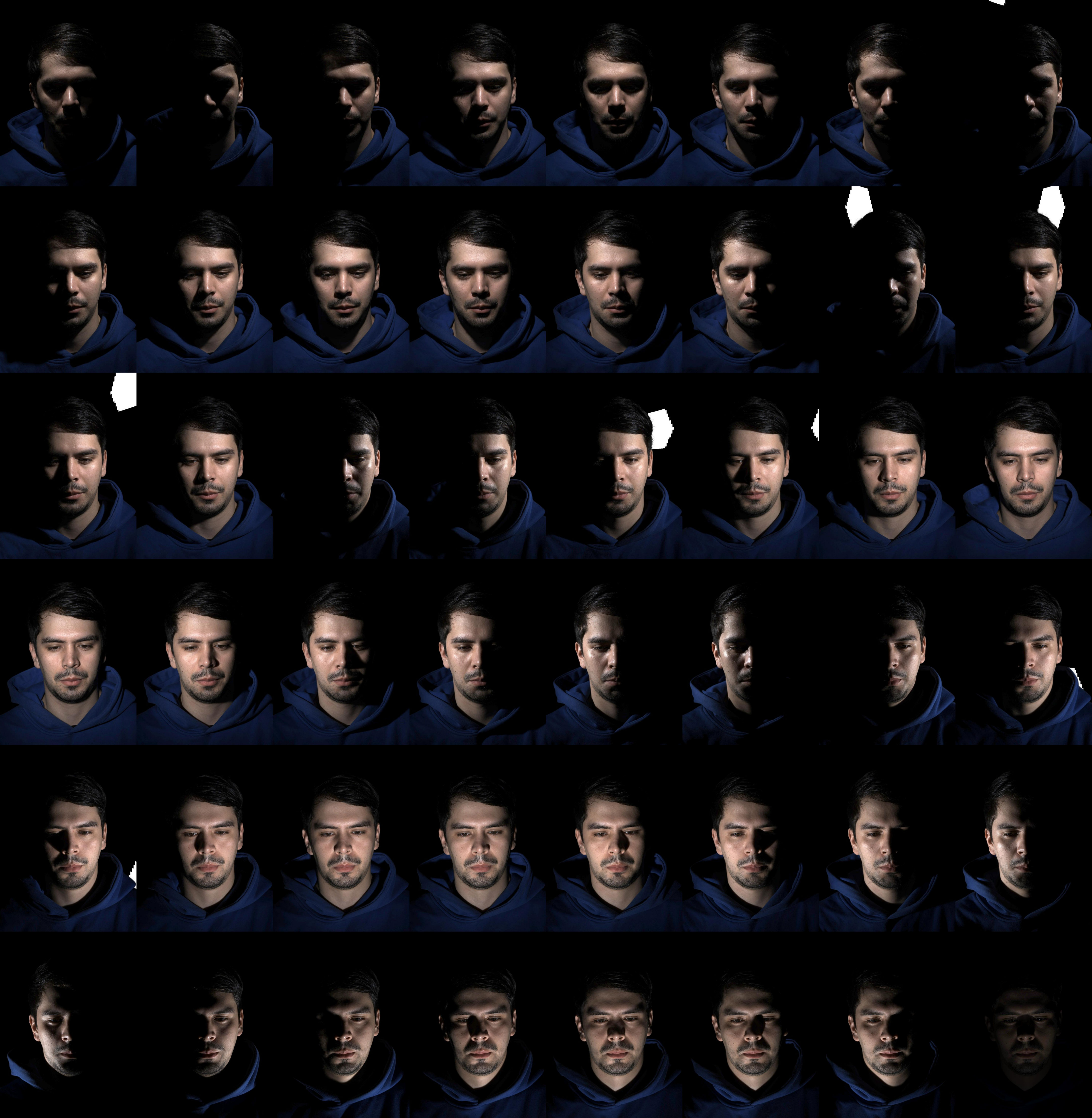}
    \caption{Real captured OLAT sequence in our POLAR dataset (selected 48 frontal LEDs).}
    \label{fig:olat}
\end{figure*}

\begin{figure*}[t]
    \centering
    \includegraphics[width=\linewidth]{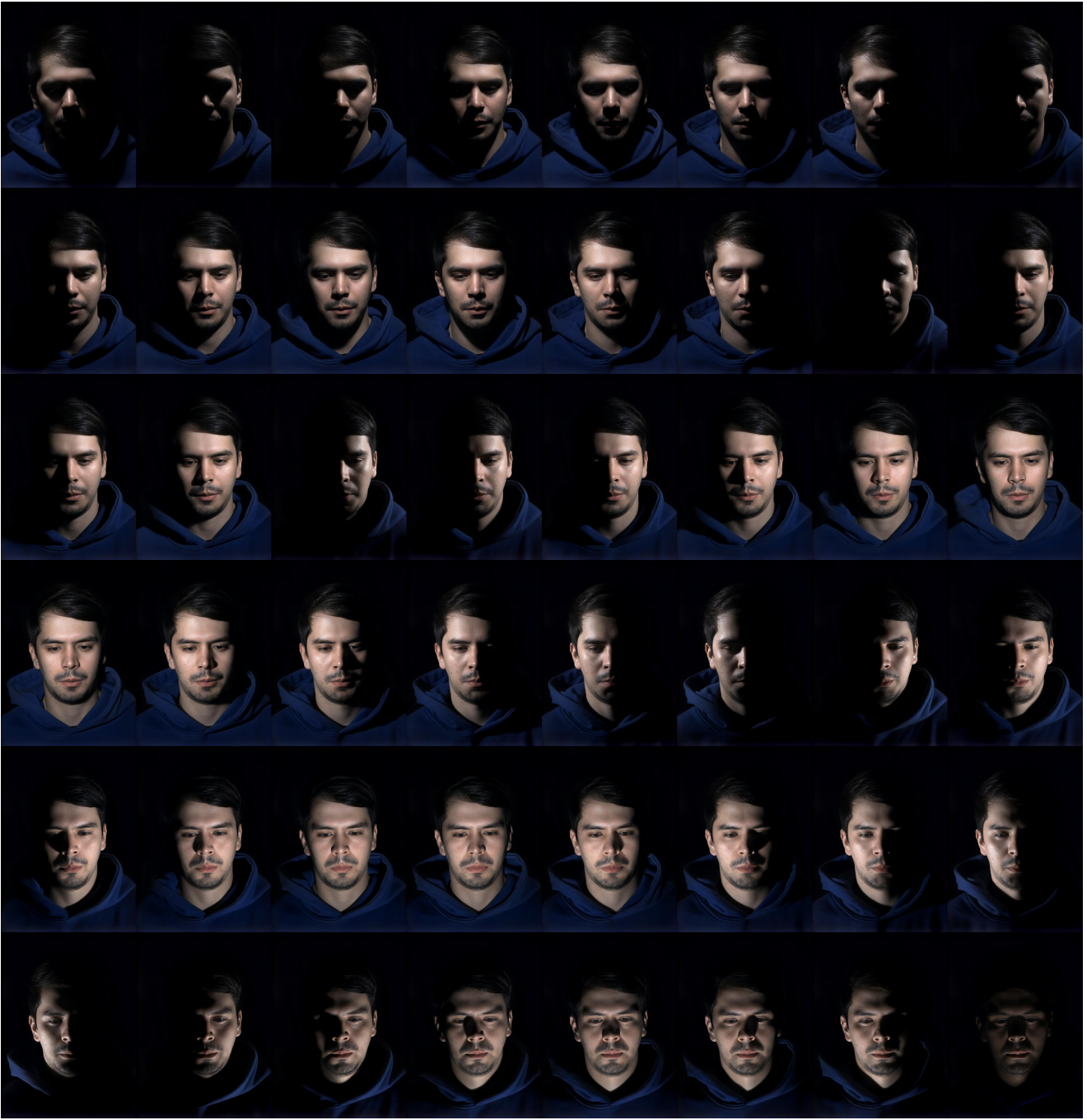}
    \caption{Generated OLAT sequence of test set by our POLARNet (selected 48 frontal LEDs).}
    \label{fig:generated_olat}
\end{figure*}

\begin{figure*}[t]
    \centering
    \includegraphics[width=\linewidth]{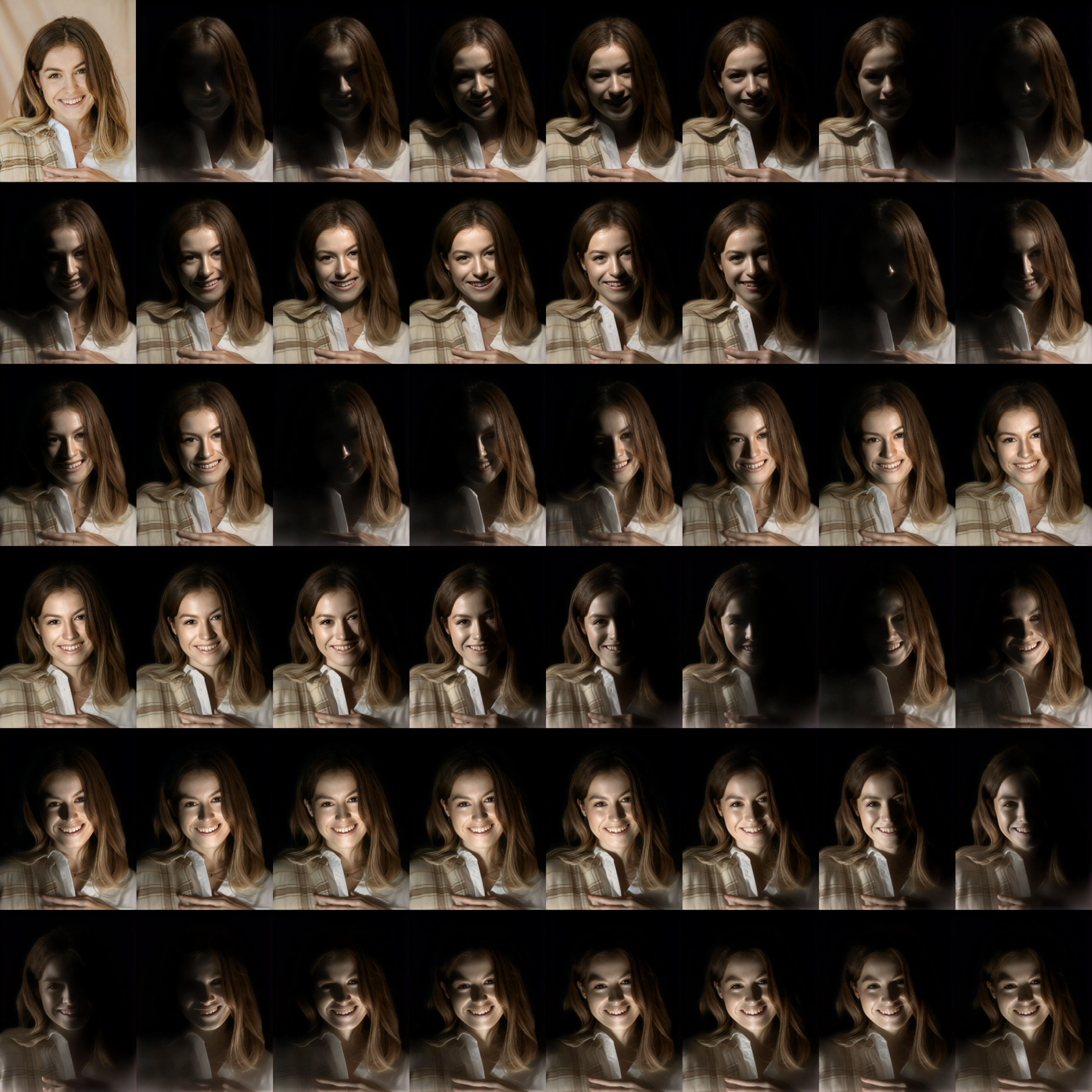}
    \caption{Generated OLAT sequence of in-the-wild portraits by our POLARNet (selected 48 frontal LEDs).}
    \label{fig:in_the_wild_olat}
\end{figure*}
\begin{figure*}[h]
    \centering
    \includegraphics[width=\linewidth]{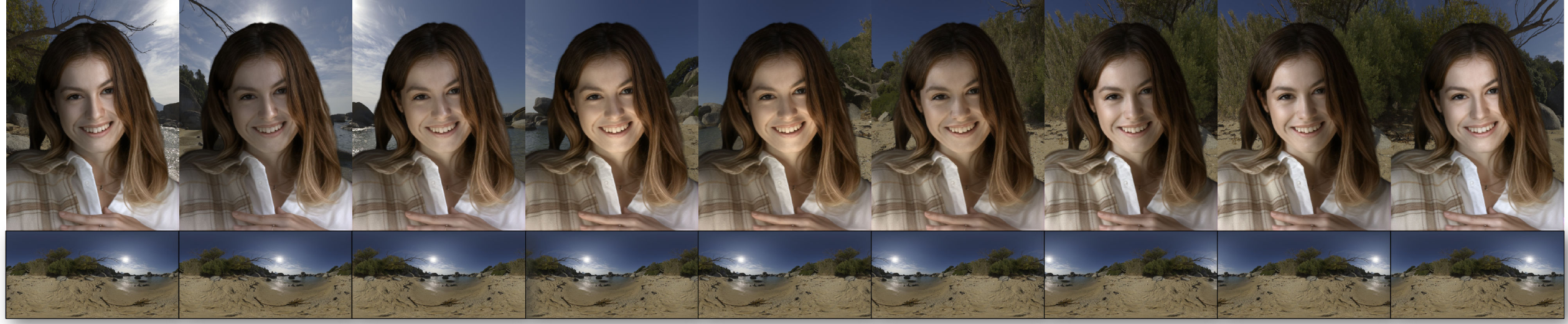}
    \caption{Relighting results under a rotating HDR map.}
    \label{fig:rotate}
\end{figure*}

\section{Failure Cases and Potential Solutions}

A typical failure case occurs when the input portrait contains strong non-uniform illumination, for example when one side of the face is significantly brighter than the other. Since POLARNet assumes a uniformly lit input image, such uneven lighting can be partially preserved in the latent representation and may propagate to the generated OLAT outputs,as shown in Fig.~\ref{fig:failure_case}. As a result, the predicted OLAT set may exhibit an unintended global shading bias that affects the quality of the synthesized relit portraits.

To address this issue, we introduce a preliminary delighting module that attempts to restore an approximately uniform-light appearance before OLAT prediction. By providing a more illumination-neutral input, the resulting OLAT estimates become more consistent across directions and the relighting quality improves accordingly. However, the current delighting module is still not fully mature and may remove subtle identity cues or fine facial details. We are exploring more robust and identity-preserving delighting strategies to further improve the reliability of the entire relighting pipeline.

\section{OLAT Visualization}
\subsection{Real Data and POLARNet Predictions}
To further demonstrate the quality of the POLAR dataset and the effectiveness of our proposed POLARNet for single-image OLAT generation, we provide a visual comparison between ground-truth OLAT captures and the corresponding OLAT predictions produced by our model. Specifically, we select 48 representative light directions that uniformly sample the frontal hemisphere and display. 
Fig.~\ref{fig:olat} presents the ground-truth OLAT samples and Fig.~\ref{fig:generated_olat} shows POLARNet predictions. 

Across all directions, POLARNet produces directionally accurate and physically meaningful illumination responses, including the movement of highlights, shading variations, and overall light–geometry interaction. The predicted OLAT images exhibit plausible specular and diffuse behavior as well as consistent energy falloff, while maintaining stable identity and global facial structure. Although some fine-scale details appear slightly smoothed, the model reliably captures the dominant lighting characteristics and avoids common artifacts such as deformation, overexposure, or inconsistent photometric shifts.

In addition to the studio examples, we also provide OLAT sequences generated from \emph{in-the-wild} portrait images in Fig.~\ref{fig:in_the_wild_olat}. It demonstrate that POLARNet generalizes beyond Light Stage conditions and is able to produce directionally consistent illumination responses on real, unconstrained images. The model preserves identity and global facial structure while delivering plausible per-light shading and highlight behaviors.

Using only a single uniform-light input, POLARNet can generate a complete set of directionally aligned, physically meaningful OLAT responses without multi-light input or multi-step diffusion sampling. These results highlight the practicality and effectiveness of our approach for controlled relighting tasks and for enabling physically interpretable illumination modeling.

\subsection{Relighting Consistency}
To further validate the physical consistency of our OLAT-based relighting pipeline, we use the generated OLAT from Fig.~\ref{fig:in_the_wild_olat} to synthesize relit results under a set of rotating outdoor environment maps, as shown in Fig.~\ref{fig:rotate}. 
As the environment map rotates, the synthesized portraits exhibit coherent and physically meaningful illumination changes. Highlights shift smoothly across facial regions, cast shadows reposition according to the dominant light direction, and the global shading pattern evolves consistently with the movement of the sun in the environment map. Importantly, POLARNet maintains stable facial identity.

\section{Implementation Details}

\paragraph{Training setup.}
We train our POLARNet on a single NVIDIA A100~(40GB) GPU using mixed-precision (FP16). 
Training uses a batch size of 4, an input resolution of $1024\times768$, and the AdamW optimizer with a learning rate of $5\times10^{-5}$. 
The model is trained for 40k steps, which takes approximately 8 hours in total. 
Our training set contains 154 subjects, covering diverse identities, facial geometry, and illumination conditions.

\paragraph{Inference performance.}
POLARNet performs single-step OLAT prediction. 
Given one uniform-light portrait, each OLAT image is generated in 0.35s, and producing the full set of 157 directions requires about 54s on a single A100 GPU.
This efficiency comes from our flow-based latent transport formulation, which avoids iterative denoising or multi-step sampling used in diffusion models.
The fast inference allows POLARNet to support interactive relighting and large-scale OLAT synthesis.


\end{document}